\documentclass{article}

\PassOptionsToPackage{numbers}{natbib}

\usepackage[preprint]{neurips_2024}
\usepackage[utf8]{inputenc} % allow utf-8 input
\usepackage[T1]{fontenc}    % use 8-bit T1 fonts
\usepackage{hyperref}       % hyperlinks
\usepackage{url}            % simple URL typesetting
\usepackage{booktabs}       % professional-quality tables
\usepackage{amsfonts}       % blackboard math symbols
\usepackage{nicefrac}       % compact symbols for 1/2, etc.
\usepackage{microtype}      % microtypography
\usepackage{xcolor}         % colors

\usepackage{mathtools} % underbrackets
\usepackage{multirow} % table multirow
\usepackage{pifont} % xmark
\newcommand{\xmark}{\ding{55}}

\title{MuDreamer: Learning Predictive World Models without Reconstruction}

\author{%
  Maxime Burchi\thanks{{\small \texttt{maxime.burchi@uni-wuerzburg.de}}}
  \quad
  Radu Timofte \\
  Computer Vision Lab, CAIDAS \& IFI, University of W\"urzburg, Germany
}

\begin{document}

\maketitle

%%%%%%%%%%%%%%%%%%%%%%%%%%%%%%%%%%%%%%%%%%%%%%%%%%%%%%%%%%%%

\begin{abstract}
The DreamerV3 agent recently demonstrated state-of-the-art performance in diverse domains, learning powerful world models in latent space using a pixel reconstruction loss. However, while the reconstruction loss is essential to Dreamer's performance, it also necessitates modeling unnecessary information. Consequently, Dreamer sometimes fails to perceive crucial elements which are necessary for task-solving when visual distractions are present in the observation, significantly limiting its potential. In this paper, we present MuDreamer, a robust reinforcement learning agent that builds upon the DreamerV3 algorithm by learning a predictive world model without the need for reconstructing input signals. Rather than relying on pixel reconstruction, hidden representations are instead learned by predicting the environment value function and previously selected actions. Similar to predictive self-supervised methods for images, we find that the use of batch normalization is crucial to prevent learning collapse. We also study the effect of KL balancing between model posterior and prior losses on convergence speed and learning stability. We evaluate MuDreamer on the commonly used DeepMind Visual Control Suite and demonstrate stronger robustness to visual distractions compared to DreamerV3 and other reconstruction-free approaches, replacing the environment background with task-irrelevant real-world videos. Our method also achieves comparable performance on the Atari100k benchmark while benefiting from faster training.
\end{abstract}
\section{Introduction}

Deep reinforcement learning has achieved great success in recent years, solving complex tasks in diverse domains. Researchers made significant progress applying advances in deep learning for learning feature representations~\cite{mnih2013playing}. The use of deep neural networks as function approximations made it possible to train powerful agents directly from high-dimensional observations like images, achieving human to superhuman performance in challenging and visually complex domains like Atari games~\cite{mnih2015human, hessel2018rainbow}, visual control~\cite{lillicrap2015continuous, barth-maron2018distributional}, the game of Go~\cite{silver2018general, schrittwieser2020mastering}, StarCraft II~\cite{vinyals2019grandmaster} and even Minecraft~\cite{baker2022video, hafner2023mastering}. However, these approaches generally require a large amount of environment interactions~\cite{tassa2018deepmind} or behavior cloning pretraining~\cite{silver2016mastering} to achieve strong performance.

To address this issue, concurrent works have chosen to focus on model-based approaches~\cite{silver2017predictron, watter2015embed}, aiming to enhance the agent performance while reducing the number of necessary interactions with the environment.
Reinforcement learning algorithms are typically categorized into two main groups: model-free algorithms, which directly learn value and/or policy functions through interaction with the environment, and model-based algorithms, which learn a model of the world. World models~\cite{sutton1991dyna, ha2018world} summarize an agent’s experience into a predictive model that can be used in place of the real environment to learn complex behaviors. This enables the agent to simulate multiple plausible trajectories in parallel, which not only enhances generalization but also improves sample efficiency.

Recent works have shown that model-based agents can effectively be trained from images, leading to enhanced performance and sample efficiency compared to model-free approaches~\cite{hafner2019learning, hafner2019dream, kaiser2019model, schrittwieser2020mastering, hafner2020mastering, ye2021mastering, micheli2022transformers}. 
The DreamerV3 agent~\cite{hafner2023mastering} recently demonstrated state-of-the-art performance across diverse domains, learning powerful world models in latent space using a pixel reconstruction loss. 
The agent solves long-horizon tasks from image inputs with both continuous and discrete action spaces. However, while the reconstruction loss is essential for Dreamer’s performance, it also necessitates modeling unnecessary information~\cite{okada2021dreaming, ma2021contrastive, nguyen2021temporal, paster2021blast, deng2022dreamerpro, bharadhwaj2022information}. Consequently, Dreamer sometimes fails to perceive crucial elements which are necessary for task-solving, significantly limiting its potential.

In this paper, we present MuDreamer, a robust reinforcement learning agent that builds upon the DreamerV3~\cite{hafner2023mastering} algorithm by learning a predictive world model without the necessity of reconstructing input signals. Taking inspiration from the MuZero~\cite{schrittwieser2020mastering} agent, MuDreamer learns a world model in latent space by predicting the environment rewards, continuation flags and value function, focusing on information relevant to the task. We also propose to incorporate an action prediction branch to predict the sequence of selected actions from the observed data. This additional task trains the world model to associate actions with environment changes and proves particularly beneficial for learning hidden representations in scenarios where environment rewards are extremely sparse. Similar to predictive self-supervised methods used for image data, we find that the use of batch normalization is crucial to prevent learning collapse in which the model produces constant or non-informative hidden states. Following~\citet{paster2021blast}, we solve this issue by introducing batch normalization inside the model representation network. We also study the effect of KL balancing between model posterior and prior losses on convergence speed and learning stability. We evaluate MuDreamer on the commonly used DeepMind Visual Control Suite~\cite{tassa2018deepmind} and demonstrate stronger robustness to visual distractions compared to DreamerV3 and other reconstruction-free approaches~\cite{nguyen2021temporal, deng2022dreamerpro}. MuDreamer learns to distinguish relevant details from unnecessary information when replacing the environment background with task-irrelevant real-world videos while DreamerV3 focuses on background details irrelevant to the task. Our method also achieves comparable performance on the Atari100k benchmark while benefiting from faster training.
\section{Related Works}

\subsection{Model-based Reinforcement Learning}

In recent years, there has been a growing interest in using neural networks as world models to simulate environments and train reinforcement learning agents from hypothetical trajectories.
Initial research primarily focused on proprioceptive tasks~\cite{gal2016improving, silver2017predictron, henaff2017model, wang2019benchmarking, wang2019exploring}, involving simple, low-dimensional environments. However, more recent efforts have shifted towards learning world models for environments with high-dimensional observations like images~\cite{kaiser2019model, hafner2019learning, hafner2019dream, schrittwieser2020mastering, ye2021mastering, micheli2022transformers}.
For example, SimPLe~\cite{kaiser2019model} successfully demonstrated planning in Atari games by training a world model in pixel space and utilizing it to train a Proximal Policy Optimization (PPO) agent~\cite{schulman2017proximal}. This approach involves a convolutional autoencoder for generating discrete latent variables and an LSTM-based recurrent network predicting latent bits autoregressively.
PlaNet~\cite{hafner2019learning} proposed to learn a Recurrent State-Space Model (RSSM) in latent space using a pixel reconstruction loss, planning using model predictive control.
Dreamer~\cite{hafner2019dream, hafner2020mastering, hafner2023mastering} extended PlaNet by incorporating actor and critic networks trained from simulated trajectories and imaginary rewards.
MuZero~\cite{schrittwieser2020mastering} took a different approach, focusing on learning a model of the environment by predicting quantities crucial for planning, such as reward, action-selection policy, and value function. This approach allowed MuZero to excel in reinforcement learning tasks without relying on the reconstruction of input observations.
EfficientZero~\cite{ye2021mastering} improved upon MuZero's sample efficiency by incorporating a representation learning objective~\cite{chen2021exploring} to achieve better performance with limited data.
Lastly, IRIS~\cite{micheli2022transformers} proposed a discrete autoencoder~\cite{van2017neural} model for imagining trajectories, predicting discrete latent tokens in an autoregressive manner, using a transformer model.

\subsection{Self-Supervised Representation Learning for Images}

Self-Supervised Learning (SSL) of image representations has attracted significant research attention in recent years for its ability to learn hidden representations from large scale unlabelled datasets. 
Reconstruction-based approaches proposed to learn hidden representations by reconstructing a corrupted version of the input image~\cite{he2022masked, xie2022simmim, feichtenhofer2022masked}. Many works focused on pixel space reconstruction while other proposed to predict hand-designed features like Histograms of Oriented Gradients (HOG)~\cite{wei2022masked}. 
Contrastive approaches proposed to learn hidden representations using joint embedding architectures where output features of a sample and its distorted version are bought close to each other, while negative samples and their distortions are pushed away~\cite{hjelm2018learning, oord2018representation, he2020momentum, chen2020simple}. These methods are commonly applied to Siamese architectures, where two identical networks are trained together, sharing parameters. 
In contrast, SwAV~\cite{caron2020unsupervised} proposed a different approach by ensuring consistency between cluster assignments produced for different augmentations of the same image, rather than directly comparing features.
Predictive approaches proposed to predict the hidden representation of a similar view of the input signal without relying on negative samples~\cite{grill2020bootstrap, caron2021emerging, ermolov2021whitening, chen2021exploring, zbontar2021barlow, baevski2022data2vec, assran2023self}. These methods prevent learning collapse using various architectural tricks such as knowledge distillation~\cite{hinton2015distilling}, normalizing output representations, or the application of additional constraints to output representations like VICReg~\cite{bardes2021vicreg}.

\subsection{Reconstruction-Free Dreamer}

Following advances in the area of self-supervised representation learning for image data, several works proposed to apply reconstruction-free representation learning techniques to Dreamer. 
The original Dreamer paper~\cite{hafner2019dream} initially experimented with contrastive learning~\cite{oord2018representation} to learn representations having maximal mutual information with the encoded observation but found that it did not match the performance of reconstruction-based representations.
Subsequently, several works proposed Dreamer variants using contrastive learning~\cite{okada2021dreaming, ma2021contrastive, nguyen2021temporal, okada2022dreamingv2, bharadhwaj2022information}, successfully competing with Dreamer on several tasks of the Visual Control Suite.
Dreaming~\cite{okada2021dreaming} proposed to use a multi-step InfoNCE loss to sequentially predict future time steps representations of augmented views. 
Temporal Predictive Coding (TPC)~\cite{nguyen2021temporal} followed a similar approach using contrastive learning to maximize the mutual information between past and the future latent states.
DreamerPro~\cite{deng2022dreamerpro} proposed to encourage uniform cluster assignment across batches of samples, implicitly pushing apart embeddings of different observations. 
Concurrently, BLAST~\cite{paster2021blast} proposed to learn hidden representation using a slow-moving teacher network to generate target embeddings~\cite{grill2020bootstrap}. BLAST also demonstrated that batch normalization was critical to the agent performance. 
In this paper, we present a reconstruction-free variant of DreamerV3 achieving comparable performance without using negatives samples, separate augmented views of images, or an additional slow-moving teacher encoder network.

\section{Background}

\subsection{Dreamer}

Our method is built on the DreamerV3~\cite{hafner2023mastering} algorithm which we refer to as Dreamer throughout the paper. Dreamer~\cite{hafner2019dream} is an actor-critic model-based reinforcement learning algorithm learning a powerful predictive world model from past experience in latent space using a replay buffer. The world model is learned from self-supervised learning by predicting the environment reward, episode continuation and next latent state given previously selected actions. The algorithm also uses a pixel reconstruction loss using an autoencoder architecture such that all information about the observations must pass through the model hidden state. The actor and critic neural networks learn behaviors purely from abstract sequences predicted by the world model. The model generates simulated trajectories from replayed experience states using the actor network to sample actions. The value network is trained to predict the sum of future reward while the actor network is trained to maximize the expected sum of future reward from the value network.

DreamerV2~\cite{hafner2020mastering} applied Dreamer to Atari games, utilizing categorical latent states with straight-through gradients~\cite{bengio2013estimating} in the world model to improve performance, instead of Gaussian latents with reparameterized gradients~\cite{kingma2013auto}. It also introduced KL balancing, separately scaling the prior cross entropy and the posterior entropy in the KL loss to encourage learning an accurate temporal prior.

DreamerV3~\cite{hafner2023mastering} mastered diverse domains using the same hyper-parameters with a set of architectural changes to stabilize learning across tasks. The agent uses Symlog predictions for the reward and value function to address the scale variance across domains. The networks also employ layer normalization~\cite{ba2016layer} to improve robustness and performance while scaling up to larger model sizes. It regularizes the policy by normalizing the returns and value function using an Exponential Moving Average (EMA) of the returns percentiles. Using these modifications, the agent solves Atari games and DeepMind Control tasks while collecting diamonds in Minecraft.

\subsection{MuZero}

MuZero~\cite{schrittwieser2020mastering} is a model-based algorithm combining Monte-Carlo Tree Search (MCTS)~\cite{coulom2006efficient} with a world model to achieve superhuman performance in precision planning tasks such as Chess, Shogi and Go. The model is learned by being unrolled recurrently for K steps and predicting environment quantities relevant to planning. All parameters of the model are trained jointly to accurately match the TD value~\cite{sutton1988learning} and reward, for every hypothetical step k. The MCTS algorithm uses the learned model to simulate environment trajectories and output an action visit distribution over the root node. This potentially better policy compared to the neural network one is used to train the policy network. MuZero excels in discrete action domains but struggles with high-dimensional continuous action spaces, where a discretization of possible actions is required to apply MCTS~\cite{ye2021mastering}. Our proposed method, MuDreamer, draws inspiration from MuZero to learn a world model by predicting the expected sum of future rewards. MuDreamer solves tasks from pixels in both continuous and discrete action space, without the need for a reconstruction loss.
\section{MuDreamer}
\label{section:method}

\begin{figure*}[ht]
        \centering
        \includegraphics[width=1.0\linewidth]{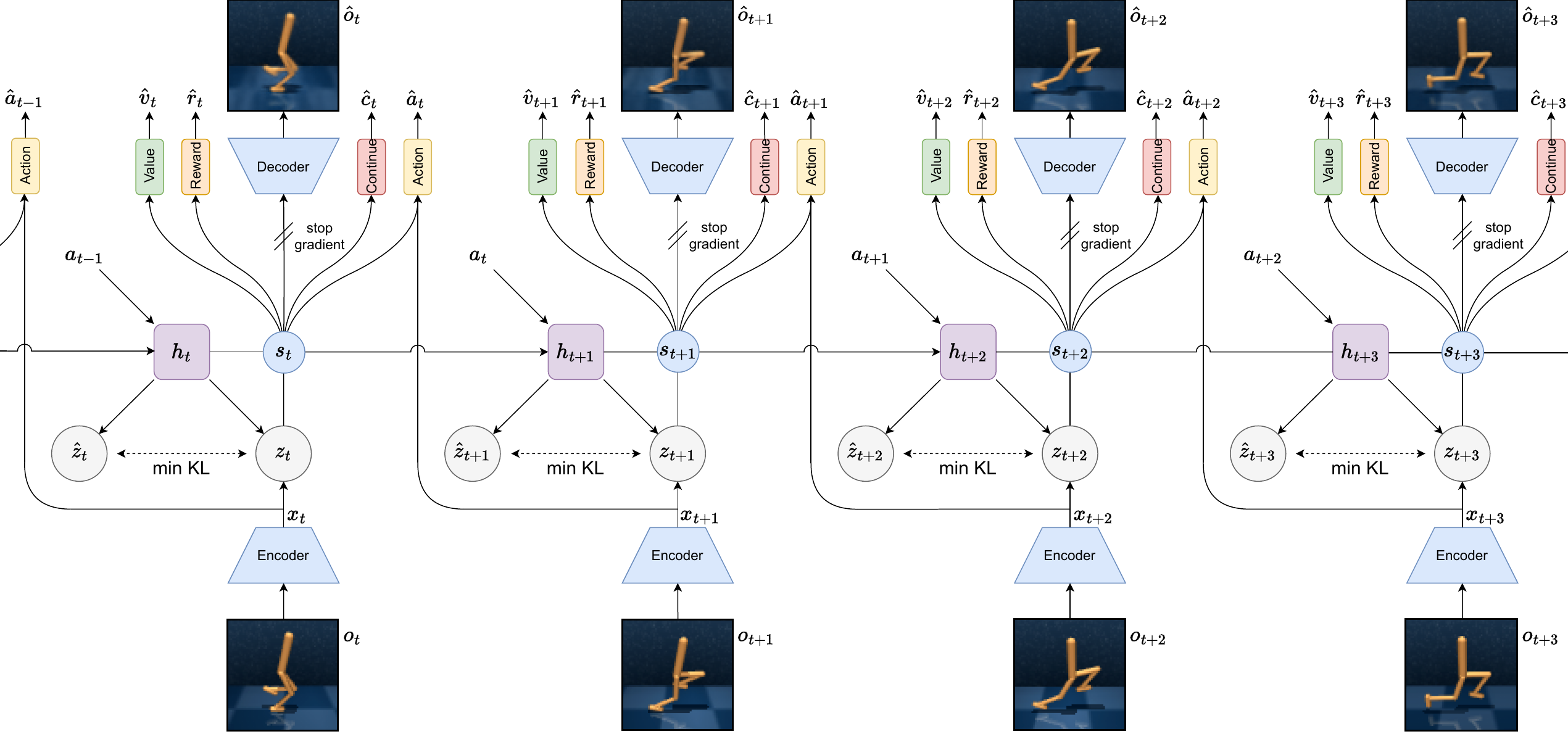}
        \caption{MuDreamer world model training. A sequence of image observations $o_{1:T}$ is sampled from the replay buffer. The sequence is mapped to hidden representations $x_{1:T}$ using a CNN encoder. At each step, the RSSM computes a posterior state $z_{t}$ representing the current observation $o_{t}$ and a prior state $\hat{z}_{t}$ that predict the posterior without having access to the current observation. Unlike Dreamer, the decoder gradients are not back-propagated to the rest of the model. The hidden representations are learned solely using value, reward, episode continuation and action prediction heads.}
        \label{figure:mu_dreamer}
\end{figure*}

We present MuDreamer, a reconstruction-free variant of the Dreamer algorithm, which learns a world model in latent space by predicting not only rewards and continuation flags but also the environment value function and previously selected actions. Figure~\ref{figure:mu_dreamer} illustrates the learning process of the MuDreamer world model. Similar to Dreamer, MuDreamer comprises three neural networks: a world model, a critic network, and an actor network. These three networks are trained concurrently using an experience replay buffer that collects past experiences. This section provides an overview of the world model and the modifications applied to the Dreamer agent. We also detail the learning process of the critic and actor networks.

\subsection{World Model Learning}

Following DreamerV3, we learn a world model in latent space, using a Convolutional Neural Network (CNN)~\cite{lecun1989backpropagation} encoder to map high-dimensional visual observations $o_{t}$ to hidden representations $x_{t}$. The world model is implemented as a Recurrent State-Space Model (RSSM)~\cite{hafner2019learning} composed of three sub networks: 
A sequential network using a Gated Recurrent Unit (GRU)~\cite{cho2014learning} to predict the next hidden state $h_{t}$ given past action $a_{t-1}$. 
A representation network predicting the current stochastic state $z_{t}$ using both $h_{t}$ and encoded features $x_{t}$. And a dynamics network predicting the stochastic state $z_{t}$ given the current recurrent state $h_{t}$. The concatenation of $h_{t}$ and $z_{t}$ forms the model hidden state $s_{t}=\{h_{t}, z_{t}\}$ from which we predict environment rewards $r_{t}$, episode continuation $c_{t} \in \{0, 1\}$ and value function $v_{t}$. We also learn an action predictor network using encoded features $x_{t}$ and preceding model hidden state $s_{t-1}$ to predict the action which led to the observed environment change. The trainable world model components are the following:

\begin{equation}
\setlength{\tabcolsep}{3pt}
\begin{tabular}{llll}
    & Encoder: & & $x_{t} = enc_{\phi}(o_{t})$  \\
    \multirow{3}{*}{RSSM $\begin{dcases*} \\ \\ \end{dcases*}$} & Sequential Network: &  & $h_{t} = f_{\phi}(h_{t-1}, z_{t-1}, a_{t-1})$    \\
    & Representation Network: & & $z_{t} \sim q_{\phi}(z_{t}\ |\ h_{t}, x_{t})$   \\
    & Dynamics Predictor: & & $\hat{z}_{t} \sim p_{\phi}(\hat{z}_{t}\ |\ h_{t})$  \\
    & Reward Predictor: & & $\hat{r}_{t} \sim p_{\phi}(\hat{r}_{t}\ |\ s_{t})$ \\
    & Continue Predictor: & & $\hat{c}_{t} \sim p_{\phi}(\hat{c}_{t}\ |\ s_{t})$ \\
    & Value Predictor: & & $\hat{v}_{t} \sim p_{\phi}(\hat{v}_{t}\ |\ s_{t})$ \\
    & Action Predictor: & & $\hat{a}_{t-1} \sim p_{\phi}(\hat{a}_{t-1}\ |\ x_{t}, s_{t-1})$ \\
    & Decoder (auxiliary): & & $\hat{o}_{t} \sim p_{\phi}(\hat{o}_{t}\ |\ sg(s_{t}))$\\
\end{tabular}
\end{equation}

Given a sequence batch of inputs $x_{1:T}$ , actions $a_{1:T}$ , rewards $r_{1:T}$ , and continuation flags $c_{1:T}$, the world model parameters ($\phi$) are optimized end-to-end to minimize a prediction loss $L_{pred}$, dynamics loss $L_{dyn}$, and representation loss $L_{rep}$ with corresponding loss weights $\beta_{pred}=1.0$, $\beta_{dyn}=0.95$ and $\beta_{rep}=0.05$. The loss function for learning the world model is:
\begin{equation}
L_{model}(\phi) = \mathrm{E}_{q_{\phi}}\Bigr[ \textstyle\sum_{t=1}^{T}(\beta_{pred}L_{pred}(\phi) + \beta_{dyn}L_{dyn}(\phi) + \beta_{rep}L_{rep}(\phi))\Bigl]
\end{equation}
The prediction loss trains the reward, continue, value and action predictors to learn hidden representations. We optionally learn an auxiliary decoder network to reconstruct the sequence of observations using the stop gradient operator $sg(.)$ to prevent the gradients from being back-propagated to other network parameters. The auxiliary decoder reconstruction loss has no effect on MuDreamer training and the decoder is excluded from total number of parameters. The dynamics loss trains the dynamics predictor to predict the next representation by minimizing the KL divergence between the predictor $p_{\phi}(\hat{z}_{t}\ |\ h_{t})$ and the next stochastic representation $q_{\phi}(z_{t}\ |\ h_{t}, x_{t})$. While the representation loss trains the representations to become more predictable if the dynamics cannot predict their distribution:
\begin{equation}
\begin{split}
L_{pred}(\phi) = & \underbracket[0.1ex]{-\log p_{\phi}(r_{t}\ |\ s_{t})}_{\text{reward predictor loss}}\ \underbracket[0.1ex]{-\log p_{\phi}(c_{t}\ |\ s_{t})}_{\text{continue predictor loss}}\ \underbracket[0.1ex]{-\log p_{\phi}(R_{t}^{\lambda} \ |\ s_{t})}_{\text{value predictor loss}} \\
& \underbracket[0.1ex]{-\log p_{\phi}(a_{t-1}\ |\ x_{t},s_{t-1})}_{\text{action predictor loss}}\ \underbracket[0.1ex]{-\log p_{\phi}(o_{t}\ |\ sg(s_{t}))}_{\text{reconstruction loss (auxiliary)}}
\end{split}
\label{equation:loss_pred}
\end{equation}
\begin{equation}
\setlength{\tabcolsep}{0pt}
\begin{tabular}{llllllllll}
    $L_{dyn}$ & $(\phi) = $ max$(1, \text{KL}[\ $ & $sg($ & $q_{\phi}(z_{t}\ |\ h_{t}, x_{t})$ &$)\ $& $||\ $ & & $p_{\phi}(\hat{z}_{t}\ |\ h_{t})$ & & $])$ \\
    $L_{rep}$ & $(\phi) = $ max$(1, \text{KL}[\ $ & & $q_{\phi}(z_{t}\ |\ h_{t}, x_{t})$& & $||\ $ & $sg($ & $p_{\phi}(\hat{z}_{t}\ |\ h_{t})$& $)\ $ & $])$ \\
\end{tabular}
\end{equation}
\textbf{Value Predictor} We use a value network to predict the environment value function from the model hidden state. The value function aims to represent the expected $\lambda$-returns~\cite{sutton1988learning} with $\lambda$ set 0.95, using a slow moving target value predictor $v_{\phi'}$ with EMA momentum $\tau=0.01$:
\begin{equation}
\setlength{\tabcolsep}{10pt}
\begin{tabular}{ll}
$R_{t}^{\lambda} = r_{t+1} + \gamma c_{t+1} \Bigl((1-\lambda)v_{\phi'}(s_{t+1}) + \lambda R_{t+1}^{\lambda}\Bigr)$ & $R_{T}^{\lambda} = v_{\phi'}(s_{T})$
\end{tabular}
\end{equation}
Following previous works~\cite{schrittwieser2020mastering, hafner2023mastering}, we set the discount factor $\gamma$ to 0.997 and use a discrete regression of $\lambda$-returns to let the critic maintain and refine a distribution over potential returns.
The $\lambda$-return targets are first transformed using the Symlog function and discretized using a \textit{twohot} encoding. Given twohot-encoded target returns $y_{t}=sg(twohot(symlog(R_{t}^{\lambda})))$ the
value predictor minimizes the categorical cross entropy loss with the predicted value logits. 

\textbf{Action Predictor} The action predictor network learns to identify the previously selected actions for each time step of the sampled trajectory. It shares the same architecture as the actor network but takes the current encoded features $x_{t}$ and preceding model hidden state $s_{t-1}$ as inputs.
Since the current model hidden state $s_{t}$ already depends of the action target $a_{t}$.

\textbf{Batch Normalization} In order to prevent learning collapse to constant or non-informative model states, like observed in predictive self-supervised learning for image data~\cite{grill2020bootstrap, chen2020simple}, we apply batch normalization~\cite{ioffe2015batch} to MuDreamer. Following BLAST~\cite{paster2021blast}, we replace the hidden normalization layer inside the representation network by a batch normalization layer. We find that it is sufficient to prevent collapse on all tasks and stabilize learning.

\begin{figure*}[ht]
        \centering
        \includegraphics[width=1.0\linewidth]{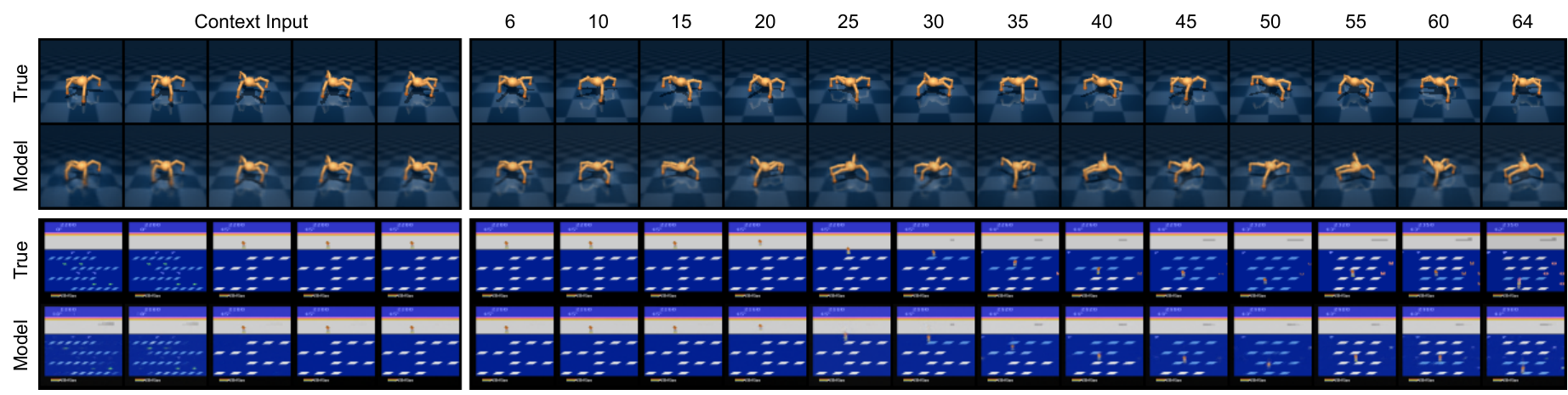}
        \caption{Reconstruction of MuDreamer model predictions over 64 time steps. We take 5 context frames and generate trajectories of 59 steps into the future using the model sequential and dynamics networks. Actions are predicted using the policy network given generated latent states. MuDreamer generates accurate long-term predictions similar to Dreamer without requiring reconstruction loss gradients during training to compress the observation information into the model hidden state.}
        \label{figure:sequence}
\end{figure*}

\subsection{Behavior Learning}

Following DreamerV3~\cite{hafner2023mastering}, MuDreamer policy and value functions are learned by imagining trajectories using the world model (Figure~\ref{figure:sequence}). Actor-critic learning takes place entirely in latent space, allowing the agent to use a large batch size of imagined trajectories. To do so, the model hidden states of the sampled sequence are flattened along batch and time dimensions. The world model imagines $H=15$ steps into the future using the sequential and dynamics networks, selecting actions from the actor. The actor and critic 
use parameter vectors ($\theta$) and ($\psi$), respectively:
\begin{equation}
\setlength{\tabcolsep}{10pt}
\begin{tabular}{llll}
    Actor: & $a_{t} \sim \pi_{\theta}(a_{t} | \hat{s}_{t})$ & Critic: & $v_{\psi}(\hat{s}_{t}) \approx \mathrm{E}_{p_{\phi}\pi_{\theta}}[\hat{R}^{\lambda}_{t}]$   
\end{tabular}
\end{equation}
\textbf{Critic Learning} Similarly to the value predictor branch, the critic is trained by predicting the discretized $\lambda$-returns, but using rewards predictions imagined by the world model:
\begin{equation}
\setlength{\tabcolsep}{10pt}
\begin{tabular}{ll}
$\hat{R}_{t}^{\lambda} = \hat{r}_{t+1} + \gamma \hat{c}_{t+1} \Bigl((1-\lambda)v_{\psi}(\hat{s}_{t+1}) + \lambda \hat{R}_{t+1}^{\lambda}\Bigr)$ & $\hat{R}_{H+1}^{\lambda} = v_{\psi}(\hat{s}_{H+1})$
\end{tabular}
\end{equation}
The critic also does not use a target network but relies on its own predictions for estimating rewards beyond the prediction horizon. This requires stabilizing the critic by adding a regularizing term of the predicted values toward the outputs of its own EMA network:
\begin{equation}
L_{critic}(\psi) = \textstyle\sum_{t=1}^{H}\bigl( \underbracket[0.1ex]{-\log p_{\psi}(\hat{R}_{t}^{\lambda} \ |\ \hat{s}_{t})}_{\text{discrete returns regression}} \underbracket[0.1ex]{-\log p_{\psi}(v_{\psi'}(\hat{s}_{t}) \ |\ \hat{s}_{t})}_{\text{critic EMA regularizer}}\ \bigr)
\end{equation}

\textbf{Actor Learning} The actor network learns to select actions that maximize advantages $A_{t}^{\lambda}=\big(\hat{R}_{t}^{\lambda} - v_{\psi}(\hat{s}_{t})\big) / \max(1, S)$ while regularizing the policy entropy to ensure exploration both in imagination and during data collection. In order to use a single regularization scale for all domains, DreamerV3 stabilizes the scale of returns using exponentially moving average statistics of their percentiles: $S=\operatorname{EMA}(\operatorname{Per}(\hat{R}_{t}^{\lambda}, 95) - \operatorname{Per}(\hat{R}_{t}^{\lambda}, 5), 0.99)$.
The actor loss computes policy gradients using stochastic back-propagation thought the model sequential and dynamics networks for continuous actions ($\rho=0$) and using reinforce~\cite{williams1992simple} for discrete actions ($\rho=1$):
\begin{equation}
\begin{aligned}
L_{actor}(\theta) & = \textstyle\sum_{t=1}^{H}\bigl(\ \underbracket[0.1ex]{-\ \rho \log \pi_{\theta}(a_{t}\ |\ \hat{s}_{t})sg(A_{t}^{\lambda})}_{\text{reinforce}} \underbracket[0.1ex]{-\ (1 - \rho)A_{t}^{\lambda}}_{\text{dynamics backprop}} \underbracket[0.1ex]{-\ \eta \mathrm{H}[\pi_{\theta}(a_{t}\ |\ \hat{s}_{t})]}_{\text{entropy regularizer}}\ \bigr)
\end{aligned}
\end{equation}
\section{Experiments}
\label{section:experiments}

In this section, we aim to evaluate the performance of our MuDreamer algorithm compared to its reconstruction-based version and other published model-based and model-free methods. We evaluate MuDreamer on the Visual Control Suite from DeepMind (Table~\ref{table:results_dmc_1M}) and under the natural background setting (Table~\ref{table:results_dmc_bg_1M}) where tasks background is replaced by real-world videos. We also proceed to a detailed ablation study, studying the effect of batch normalization, action-value predictors and KL balancing on performance and learning stability (Table~\ref{table:ablation_dmc}). The performance curves showing comparison for each individual task, Atari100k benchmark comparison with other model-based methods, as well as additional visualizations can be found in the appendix.

\subsection{Visual Control Suite}
\label{subsection:visual_control_suite}

The DeepMind Control Suite was introduced by~\citet{tassa2018deepmind} as a set of continuous control tasks with a standardised structure and interpretable rewards. The suite is intended to serve as performance benchmarks for reinforcement learning agents in continuous action space. The tasks can be solved from low-dimensional inputs and/or pixel observations. In this work, we evaluate our method on the Visual Control Suite benchmark which contains 20 tasks where the agent receives only high-dimensional images as inputs and a budget of 1M environment steps. Similarly to DreamerV3~\cite{hafner2023mastering}, we use 4 environment instances during training with a training ratio of 512 replayed steps per policy step. Our PyTorch~\cite{paszke2019pytorch} implementation of MuDreamer takes a little less than 14 hours to reach 1M environment steps on the \textit{Walker Run} visual control task using a single NVIDIA RTX 3090 GPU and 16 CPU cores, while DreamerV3 takes 15 hours. The total amount of trainable parameters for MuDreamer and DreamerV3 are 15.3M and 17.9M, respectively.

\begin{table*}[ht]
    \caption{Visual Control Suite scores (1M environment steps). $\dagger$~results were taken from~\citet{hafner2023mastering}. We average the evaluation score over 10 episodes and use 3 different seeds per experiment.} 
    \setlength{\tabcolsep}{10pt}
    \scriptsize
    \centering
    \begin{tabular}{lccccc}
        \toprule
        Task & SAC$^{\dagger}$ & CURL$^{\dagger}$ & DrQ-v2$^{\dagger}$ & DreamerV3$^{\dagger}$ & MuDreamer \\ 
        \midrule
        Acrobot Swingup & 5.1 & 5.1 & 128.4 & 210.0 & \textbf{304.6} \\
        Cartpole Balance & 963.1 & 979.0 & 991.5 & \textbf{996.4} & 990.4 \\
        Cartpole Balance Sparse & 950.8 & 981.0 & 996.2 & \textbf{1000.0} & \textbf{1000.0} \\
        Cartpole Swingup & 692.1 & 762.7 & \textbf{858.9} & 819.1 & 823.0 \\
        Cartpole Swingup Sparse & 154.6 & 236.2 & 706.9 & \textbf{792.9} & 582.0 \\
        Cheetah Run & 27.2 & 474.3 & 691.0 & 728.7 & \textbf{872.5} \\
        Cup Catch & 163.9 & \textbf{965.5} & 931.8 & 957.1 & 930.8 \\
        Finger Spin & 312.2 & \textbf{877.1} & 846.7 & 818.5 & 603.6 \\
        Finger Turn Easy & 176.7 & 338.0 & 448.4 & 787.7 & \textbf{915.4} \\
        Finger Turn Hard & 70.5 & 215.6 & 220.0 & 810.8 & \textbf{886.5} \\
        Hopper Hop & 3.1 & 152.5 & 189.9 & \textbf{369.6} & 311.8 \\
        Hopper Stand & 5.2 & 786.8 & 893.0 & \textbf{900.6} & 883.9 \\
        Pendulum Swingup & 560.1 & 376.4 & \textbf{839.7} & 806.3 & 806.7 \\
        Quadruped Run & 50.5 & 141.5 & 407.0 & 352.3 & \textbf{627.8} \\
        Quadruped Walk & 49.7 & 123.7 & 660.3 & 352.6 & \textbf{860.0} \\
        Reacher Easy & 86.5 & 609.3 & 910.2 & 898.9 & \textbf{907.0} \\
        Reacher Hard & 9.1 & 400.2 & 572.9 & 499.2 & \textbf{733.0} \\
        Walker Run & 26.9 & 376.2 & 517.1 & \textbf{757.8} & 740.9 \\
        Walker Stand & 159.3 & 463.5 & 974.1 & \textbf{976.7} & 964.3 \\
        Walker Walk & 38.9 & 828.8 & 762.9 & \textbf{955.8} & 949.8 \\
        \midrule
        Mean & 225.3 & 504.7 & 677.4  & 739.6 & \textbf{784.7} \\
        Median & 78.5  &  431.8 & 734.9 & 808.5 & \textbf{849.6} \\
        \bottomrule
    \end{tabular}
    \label{table:results_dmc_1M}
\end{table*}

Table~\ref{table:results_dmc_1M} compares MuDreamer with DreamerV3 and other recent methods on the Visual Control Suite using 1M environment steps. MuDreamer achieves state-of-the-art mean-score without reconstructing the input signal, outperforming SAC~\cite{haarnoja2018soft}, CURL~\cite{laskin2020curl}, DrQ-v2~\cite{yarats2021mastering} and DreamerV3. 
As shown in Figure~\ref{figure:sequence}, although the reconstruction gradients are not propagated to the whole network, the decoder easily reconstruct the input image, meaning that the model hidden state contains all necessary information about the environment. We find that MuDreamer achieves better performance and learning stability on tasks like \textit{Finger Turn Hard} and \textit{Reacher Hard} where crucial task-solving elements are small and harder to model using reconstruction. 

\subsection{Natural Background Setting}

\begin{figure}[!ht]
        \centering
        \includegraphics[width=0.95\linewidth]{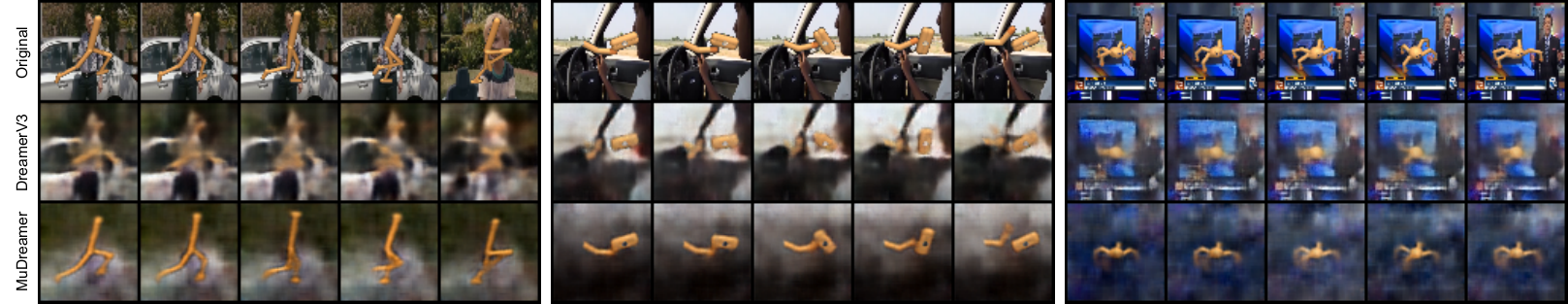}
        \caption{Agents reconstruction of observations using natural backgrounds for \textit{Walker Run}, \textit{Finger Spin} and \textit{Quadruped Run} tasks. First row shows original sequence of observation, second row shows DreamerV3 reconstruction and third row MuDreamer decoder reconstruction. We observe that DreamerV3 reconstructs general details while MuDreamer learns to filter unnecessary information.}
        \label{figure:observation_dmc_background}
\end{figure}

While the reconstruction loss is essential for Dreamer’s performance, it also necessitates modeling unnecessary information. Consequently, Dreamer sometimes fails to perceives crucial elements which are necessary for task-solving when visual distractions are present in the observation, significantly limiting its potential. This object vanishing phenomenon arises particularly when crucial elements are small like the ball in \textit{Cup Catch}. In order to further study the effect of visual distractions on DreamerV3 and MuDreamer performance, we experiment with the natural background setting~\cite{ma2021contrastive, nguyen2021temporal, deng2022dreamerpro, bharadhwaj2022information}, where the DeepMind Visual Control tasks background is replaced by real-world videos. Following TPC~\cite{nguyen2021temporal} and DreamerPro~\cite{deng2022dreamerpro}, we replace the background by task-irrelevant videos randomly sampled from the \textit{driving car} class of the Kinetics 400 dataset~\cite{kay2017kinetics}. We also use two separate sets of background videos for training and evaluation in order to test generalization to unseen distractions. Table~\ref{table:results_dmc_bg_1M} shows the comparison of MuDreamer with DreamerV3, DreamerPro and TPC on the Visual Control Suite under the natural background setting. MuDreamer successfully learns a policy on most tasks and achieves state-of-the-art performance with a mean-score of 517.0 compared to 445.2 and 372.8 for DreamerPro and TPC. Figure~\ref{figure:observation_dmc_background} shows the decoder reconstruction of observations by DreamerV3 and MuDreamer on \textit{Walker Run}, \textit{Finger Spin} and \textit{Quadruped Run} tasks. MuDreamer correctly reconstructs the agent body with a monochrome or blurry background while DreamerV3 focuses on the background details, discarding the agent body and necessary information.

\begin{table}[ht]
    \caption{Visual Control Suite scores under the natural background setting (1M env. steps). $\dagger$ denotes our tested reimplementation of DreamerV3. $\ddagger$ results were obtained using the official implementation of DreamerPro. We average the evaluation score over 10 episodes and use 3 seeds per experiment.} 
    \setlength{\tabcolsep}{10pt}
    \scriptsize
    \centering
    \hfill \break
    \begin{tabular}{lccccc}
    \toprule
    Task & Random & TPC$^{\ddagger}$ & DreamerPro$^{\ddagger}$ & DreamerV3$^{\dagger}$ & MuDreamer \\ 
    \midrule
    Acrobot Swingup & 0.3 & 5.1 & 13.1 & 9.1 & \textbf{41.9} \\
    Cartpole Balance & 329.3 & 792.9 & 870.1 & 198.7 & \textbf{974.8} \\
    Cartpole Balance Sparse & 53.9 & 26.9 & 198.4 & 18.4 & \textbf{898.7} \\
    Cartpole Swingup & 67.4 & 574.8 & 689.2 & 145.7 & \textbf{794.4} \\
    Cartpole Swingup Sparse & 0.0 & 0.2 & \textbf{17.8} & 0.3 & 0.0 \\
    Cheetah Run & 6.7 & 440.8 & \textbf{380.7} & 94.3 & 318.1 \\
    Cup Catch & 31.5 & 451.5 & 437.5 & 27.9  & \textbf{904.5} \\
    Finger Spin & 0.9 & 696.8 & \textbf{724.2} & 96.5 & 644.2 \\
    Finger Turn Easy & 48.8 & \textbf{479.5} & 232.4 & 197.8 & 229.4 \\
    Finger Turn Hard & 35.0 & 198.3 & \textbf{228.3} & 39.8 & 226.7 \\
    Hopper Hop & 0.0 & 0.2 & \textbf{1.4} & 0.6 & 0.2 \\
    Hopper Stand & 1.9 & 14.5 & \textbf{296.5} & 3.0 & 5.4 \\
    Pendulum Swingup & 2.0 & \textbf{778.7} & 777.6 & 8.0 & 606.8 \\
    Quadruped Run & 8.8 & 162.9 & 470.8 & 108.9 & \textbf{735.0} \\
    Quadruped Walk & 110.0 & 681.4 & 784.5 & 61.2 & \textbf{872.8} \\
    Reacher Easy & 52.6 & 642.4 & 692.7 & 154.2 & \textbf{914.4} \\
    Reacher Hard & 7.4 & 7.0 & 9.4 & 10.6 & \textbf{13.5} \\
    Walker Run & 25.9 & 137.9 & 402.9 & 78.7 & \textbf{432.8} \\
    Walker Stand & 139.4 & 935.4 & 940.6 & 254.4 & \textbf{966.7} \\
    Walker Walk & 36.8 & 428.3 & 736.1 & 164.7 & \textbf{759.0} \\
    \midrule
    Mean  & 47.9 & 372.8 & 445.2 & 83.6 & \textbf{517.0} \\
    Median  & 28.7 & 409.8 & 416.2 & 57.1 & \textbf{620.0} \\
    \bottomrule
    \end{tabular}
    \label{table:results_dmc_bg_1M}
\end{table}

\subsection{Ablation Study}

\begin{figure*}[ht] 
    \centering
    \includegraphics[width=0.9\textwidth]{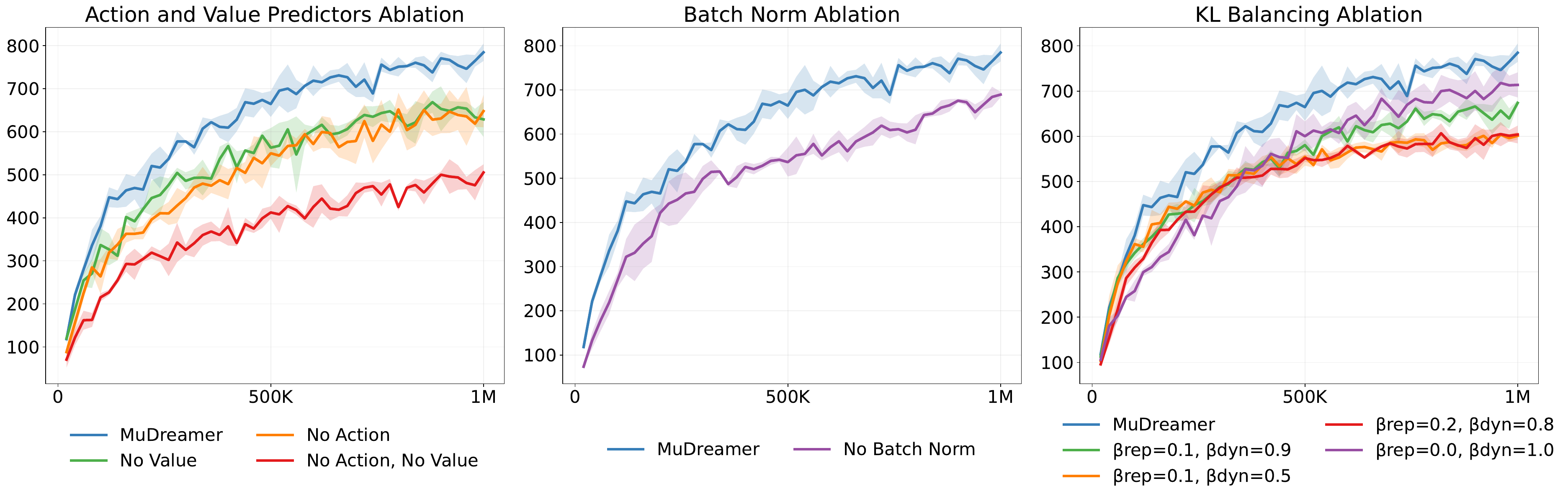}
    \caption{Ablations mean scores on the Visual Control Suite using 1M environment steps.}
    \label{figure:ablation_dmc}
\end{figure*}

In order to understand the necessary components of MuDreamer, we conduct ablation studies applying one modification at a time. We study the impact of using value and action prediction branches, removing one or both branches during training. We study the effect of using batch normalization in the representation network to stabilize learning and avoid collapse without the reconstruction loss. We also study the effect of KL balancing hyper-parameters on learning speed and stability. We perform all these ablation studies on the Visual Control Suite, observing the effect of each modification for diverse tasks. The mean and median score results of these studies are summarized in Table~\ref{table:ablation_dmc} and Figure~\ref{figure:ablation_dmc}. Please refer to the appendix for the evaluation curves of individual tasks.

\textbf{Action-Value Predictors} We study the necessity of using action and value prediction branches to learn hidden representations and successfully solve the tasks without reconstruction loss. We observed that removing the action or value prediction heads led to a degradation of MuDreamer performance and decoder reconstruction quality on most of the Visual Control tasks. The action and value prediction losses require the model to learn environment dynamics which leads to more accurate model predictions and better performance.

\begin{table}[ht]
    \caption{Ablations evaluated on 20 tasks of the Visual Control Suite using 1M environment steps. Each ablation applies only one modification to the MuDreamer agent.}
    \setlength{\tabcolsep}{15pt}
    \scriptsize
    \centering
    \hfill \break
    \begin{tabular}{lccc}
    \toprule
    Agent & Mean & Median & Task Score $\geq$ MuDreamer \\ 
    \midrule
    MuDreamer & \textbf{784.7} & \textbf{849.6} &  \\
    No Value Predictor & 628.8 & 745.6 &  5 / 20 \\
    No Action Predictor & 648.0 & 810.8 & 6 / 20 \\
    No Action and Value Predictors & 505.4 & 493.2 & 4 / 20\\
    No Batch Normalization & 689.5 & 820.7 & 8 / 20 \\
    $\beta_{rep}=0.0$, $\beta_{dyn}=1.0$ & 713.5 & 824.3 & 6 / 20\\
    $\beta_{rep}=0.1$, $\beta_{dyn}=0.9$ & 673.7 & 845.7 & 8 / 20 \\
    $\beta_{rep}=0.1$, $\beta_{dyn}=0.5$ & 601.1 & 790.2 & 10 / 20\\
    $\beta_{rep}=0.2$, $\beta_{dyn}=0.8$ & 603.9 & 694.5 & 5 / 20 \\
    \bottomrule
    \end{tabular}
    \label{table:ablation_dmc}
\end{table}

\textbf{Batch Normalization} We study the effect of using batch normalization instead of layer normalization inside the representation network to stabilize representation learning. We find that batch normalization prevents collapse in which the model produces constant or non-informative hidden states. Without batch normalization, we observe that MuDreamer fails to learn representations for some of the tasks. The standard deviation of the model encoded features $x_{t}$ converges to zero, which prevent learning hidden representations. Batch normalization solves this problem by stabilizing dynamics and representation losses.

\textbf{KL Balancing} We find that applying the default KL balancing parameters of DreamerV3 ($\beta_{dyn}=0.5$, $\beta_{rep}=0.1$) slows down convergence for some of the tasks, restraining the model from learning representations. A larger regularization of the posterior toward the prior with $\beta_{rep}=0.2$ limits the amount of information encoded in the model stochastic state. We observe that unnecessary information such as the environment floor and the agent shadow are sometimes reconstructed as monochrome surfaces without the original details. Similar to BLAST and predictive SSL approaches~\cite{grill2020bootstrap, chen2020simple}, we experiment using the stop gradient operation with $\beta_{rep}=0.0$. We find that this solves the issue but generates learning instabilities with spikes for the dynamics and prediction losses. We suppose this create difficulties for the prior to predict posterior representations since we do not use a slow moving teacher encoder network. Using a slight regularization of the representations with $\beta_{rep}=0.05$ solved both of these issues.
\section{Conclusion}
We presented MuDreamer, a robust reinforcement learning agent solving control tasks from image inputs with continuous and discrete action spaces, all without the need to reconstruct input signals. MuDreamer learns a world model by predicting environment rewards, value function, and continuation flags, focusing on information relevant to the task. We also proposed to incorporate an action prediction branch to predict the sequence of selected actions. MuDreamer demonstrates stronger robustness to visual distractions on the DeepMind Visual Control Suite compared to DreamerV3 and other methods when the environment background is replaced by task-irrelevant real-world videos. Our approach also benefits from faster training, as it does not require training an additional decoder network or encoding separate augmented views of the observation to learn hidden representations.

%%%%%%%%%%%%%%%%%%%%%%%%%%%%%%%%%%%%%%%%%%%%%%%%%%%%%%%%%%%%

\bibliographystyle{plainnat}
\bibliography{references}
%%%%%%%%%%%%%%%%%%%%%%%%%%%%%%%%%%%%%%%%%%%%%%%%%%%%%%%%%%%%

\clearpage
\appendix
\section{Appendix}

\subsection{Atari100k Benchmark}
\label{subsection:atari100k_benchmark}

The Atari100k benchmark was proposed in~\citet{kaiser2019model} to evaluate reinforcement learning agents on Atari games in low data regime. The benchmark includes 26 Atari games with discrete actions and a budget of 400K environment steps, amounting to 100K policy steps using the default action repeat setting. This represents 2 hours of real-time play. The current state-of-the-art is held by EfficientZero (EffZero)~\cite{ye2021mastering}, which uses look-ahead search to select the best action, with a human mean score of 190\% and median of 116\%. In this work, we compare our method with DreamerV3 and other model-based approaches which do not use look-ahead search.

\begin{table}[!ht]
    \caption{Atari100k scores (400K environment steps).  We average the evaluation score over 10 episodes and use 5 different seeds per experiment.}
    \centering
    \scriptsize
    \hfill \break
    \begin{tabular}{lcccccccc}
    \toprule
     &  &  & \multicolumn{2}{c}{Lookahead search} & \multicolumn{4}{c}{No lookahead search} \\
    \cmidrule(lr){4-5}\cmidrule(lr){6-9} Game & Random & Human & MuZero & EffZero & SimPLe & IRIS & DreamerV3 & MuDreamer \\ 
    \midrule
    Alien & 228 & 7128 & 530 & 1140 & 617 & 420 & \textbf{959} & 951 \\
    Amidar & 6 & 1720 & 39 & 102 & 74 & 143 & 139 & \textbf{153} \\
    Assault & 222 & 742 & 500 & 1407 & 527 & \textbf{1524} & 706 & 891 \\
    Asterix & 210 & 8503 & 1734 & 16844 & 1128 & 854 & 932 & \textbf{1411} \\
    Bank Heist & 14 & 753 & 193 & 362 & 34 & 53 & \textbf{649} & 156 \\
    Battle Zone & 2360 & 37188 & 7688 & 17938 & 4031 & \textbf{13074} & 12250 & 12080 \\
    Boxing & 0 & 12 & 15 & 44 & 8 & 70 & 78 & \textbf{96} \\
    Breakout & 2 & 30 & 48 & 406 & 16 & \textbf{84} & 31 & 34 \\
    Chopper Com. & 811 & 7388 & 1350 & 1794 & 979 & \textbf{1565} & 420 & 808 \\
    Crazy Climber & 10780 & 35829 & 56937 & 80125 & 62584 & 59324 & \textbf{97190} & 96128  \\
    Demon Attack & 152 & 1971 & 3527 & 13298 & 208 & \textbf{2034} & 303 & 553 \\
    Freeway & 0 & 30 & 22 & 22 & 17 & \textbf{31} & 0 & 5  \\
    Frostbite & 65 & 4335 & 255 & 314 & 237 & 259 & 909 & \textbf{1652}  \\
    Gopher & 258 & 2412 & 1256 & 3518 & 597 & 2236 & \textbf{3730} & 1500 \\
    Hero & 1027 & 30826 & 3095 & 8530 & 2657 & 7037 & \textbf{11161} & 8272 \\
    James Bond & 29 & 303 & 88 & 459 & 100 & \textbf{463} & 445 & 409 \\
    Kangaroo & 52 & 3035 & 63 & 962 & 51 & 838 & 4098 & \textbf{4380} \\
    Krull & 1598 & 2666 & 4891 & 6047 & 2205 & 6616 & 7782 & \textbf{9644} \\
    Kung Fu Mas. & 258 & 22736 & 18813 & 31112 & 14862 & 21760 & 21420 & \textbf{26832} \\
    Ms Pacman & 307 & 6952 & 1266 & 1387 & 1480 & 999 & 1327 & \textbf{2311} \\
    Pong & –21 & 15 & -7 & 21 & 13 & 15 & \textbf{18} & \textbf{18} \\
    Private Eye & 25 & 69571 & 56 & 100 & 35 & 100 & 882 & \textbf{1042} \\ 
    Qbert & 164 & 13455 & 3952 & 15458 & 1289 & 746 & 3405 & \textbf{4061} \\
    Road Runner & 12 & 7845 & 2500 & 18512 & 5641 & 9615 & \textbf{15565} & 8460 \\ 
    Seaquest & 68 & 42055 & 208 & 1020 & \textbf{683} & 661 & 618 & 428 \\
    Up N Down & 533 & 11693 & 2897 & 16096 & 3350 & 3546 & 7600 & \textbf{26494} \\
    \midrule
    \#Superhuman & 0 & N/A & 5 & 13 & 1 & 10 & 9 & \textbf{11} \\
    Human Mean & 0\% & 100\% & 56\% & 190\% & 33\%  & 105\% & 112\% & \textbf{126\%} \\
    Human Median & 0\% & 100\% & 23\% & 116\% & 13\% & 29\% & \textbf{49\%} & 43\% \\
    \bottomrule
    \end{tabular}
    \label{table:results_atari100k}
\end{table}

Table~\ref{table:results_atari100k} compares MuDreamer with DreamerV3~\cite{hafner2023mastering} and other model-based approaches~\cite{schrittwieser2020mastering, chen2020simple, laskin2020curl, ye2021mastering, micheli2022transformers} on the Atari100k benchmark. Following preceding works, we compare the mean and median returns across all 26 games using human-normalized scores calculated as follow: 
\begin{equation}
    \texttt{norm\_score}=(\texttt{agent\_score} - \texttt{random\_score})~/~(\texttt{human\_score} - \texttt{random\_score})
\end{equation}
MuDreamer achieves results comparable to DreamerV3, without using a decoder reconstruction loss during training. Our implementation takes 4 hours and 40 minutes to reach 400K steps on the \textit{Breakout} Atari game using a single NVIDIA RTX 3090 GPU and 16 CPU cores, while DreamerV3 takes 5 hours and 20 minutes. 

\subsection{Limitations}
\label{section:limitations}

% Limitations of our work include the fact that while MuDreamer achieves state-of-the-art performance on the natural background setting, it still does not successfully solve all tasks. MuDreamer fails to solves more complicated tasks like \textit{Acrobot Swingup} and \textit{Hopper Hop} when the background is replaced with real-world videos. 
Certain limitations of our work arise from the fact that trajectories sampled from the replay buffer are off-policy, which requires the value predictor to fit the value of older policies that are no longer accurate. The resulting value function differs from the one learned by the critic network which fits the value of the current neural network policy thought the world model. While we found this to have no effect on performance, experimenting with off-policy correction~\cite{ye2021mastering}, this may be the case when scaling up to larger model sizes and more complex tasks.

\newpage

\subsection{Ethics Statement}
\label{section:ethics}

The development of autonomous agents for real-world applications introduces various safety and environmental concerns. In real-world scenarios, an agent might cause harm to individuals and damage to its surroundings during both training and deployment. Although using world models during training can mitigate these risks by allowing policies to be learned through simulation, some risks may still persist. This statement aims to inform users of these potential risks and emphasize the importance of AI safety in the application of autonomous agents to real-world scenarios.

\subsection{MuDreamer Hyper-Parameters}
\label{subsection:hyper-parameters}

\begin{table*}[!ht]
\caption{MuDreamer hyper-parameters applied to the DeepMind Visual Control Suite (DMC) and Atari100k benchmark. We apply the same hyper-parameters to all tasks except for the number of environment instances and the training ratio (number of replayed steps per policy step). Discount and Return Lambda values are kept the same for World Model and Actor Critic learning.}
\scriptsize
\setlength{\tabcolsep}{20pt}
\renewcommand{\arraystretch}{1.25}
\centering
\hfill \break
\begin{tabular}{lcc}
\toprule
Parameter & Symbol & Setting\\ 
\midrule
General & &\\
Replay Buffer Capacity (FIFO) & --- & $10^{6}$ \\
Batch Size & B & 16 \\
Batch Length & T & 64 \\
Optimizer & --- & Adam \\
Activation & --- & LayerNorm + SiLU \\
Model Size & --- & DreamerV3 Small \\ 
Input Image Resolution & --- & $64 \times 64$ RGB \\
Replayed Steps per Policy Step & --- & 512 (DMC) / 1024 (Atari100k) \\
Environment Instances & --- & 4 (DMC) / 1 (Atari100k) \\
\midrule
World Model & & \\
Number of Latents & --- & 32 \\
Classes per Latent & --- & 32 \\
Prediction Loss Scale & $\beta_{pred}$ & 1.0 \\
Dynamics Loss Scale & $\beta_{dyn}$ & 0.95 \\
Representation Loss Scale & $\beta_{rep}$ & 0.05 \\
Learning Rate & $\alpha$ & $10^{-4}$ \\
Adam Betas & $\beta_{1}$, $\beta_{2}$ & 0.9, 0.999 \\
Adam Epsilon & $\epsilon$ & $10^{-8}$ \\
Gradient Clipping & --- & 1000 \\
Slow Value Momentum & $\tau$ & 0.99 \\
Model Discount & $\gamma$ & 0.997 \\
Model Return Lambda & $\lambda$ & 0.95 \\
Activation Representation Network & --- & BatchNorm + SiLU \\
BatchNorm Momentum & --- & 0.9 \\
\midrule
Actor Critic & & \\
Imagination Horizon & H & 15 \\
Discount & $\gamma$ & 0.997 \\
Return Lambda & $\lambda$ & 0.95 \\
Critic EMA Decay & --- & 0.98 \\
Critic EMA regularizer & --- & 1.0 \\
Return Normalization Percentiles & --- & $5^{th}$ and $95^{th}$ \\
Return Normalization Decay & --- & 0.99 \\
Actor Entropy Scale & $\eta$ & $3 \cdot 10^{-4}$ \\
Learning Rate & $\alpha$ & $3 \cdot 10^{-5}$ \\
Adam Betas & $\beta_{1}$, $\beta_{2}$ & 0.9, 0.999 \\
Adam Epsilon & $\epsilon$ & $10^{-5}$ \\
Gradient Clipping & --- & 100 \\
\bottomrule
\end{tabular}
\label{table:hyperparams}
\end{table*}

\newpage

\subsection{Network Architecture Details}

\begin{table*}[!ht]
\caption{Encoder network architecture. LN refers to Layer Normalization~\cite{ba2016layer}. The encoder downsamples images with strided convolutions of kernel size $=4$, stride $=2$ and zero padding $=1$.}
\scriptsize
\centering
\hfill \break
\begin{tabular}{cc}
\toprule
Module & Output shape \\
\midrule
Input image ($o_{t}$) & $3 \times 64 \times 64$ \\
Conv2d + LN + SiLU & $32 \times 32 \times 32$ \\
Conv2d + LN + SiLU & $64 \times 16 \times 16$ \\
Conv2d + LN + SiLU & $128 \times 8 \times 8$ \\
Conv2d + LN + SiLU & $256 \times 4 \times 4$ \\
Flatten and Output features ($x_{t}$) & $4096$ \\
\bottomrule
\end{tabular}
\label{table:encoder}
\end{table*}

\begin{table*}[!ht]
\caption{The Sequential Network process previous stochastic state $z_{t-1}$ and action $a_{t-1}$ to compute the next model hidden state $h_{t}$ using a Gated Recurrent Unit (GRU)~\cite{cho2014learning}. The GRU conditions the model hidden state on past context $h_{t-1}$ which helps the world model to make accurate predictions. The initial GRU hidden state vector $h_{0}$ is learned via gradient descent among other network parameters and processed by the dynamics predictor to generate the initial stochastic state $z_{0}$.}
\scriptsize
\centering
\hfill \break
\begin{tabular}{cc}
\toprule
Module & Output shape \\
\midrule
Input stochastic state ($z_{t-1}$), Input action ($a_{t-1})$ & $32 \times 32$, $A$ \\
Flatten and Concatenate & $1024 + A$ \\
Linear + LN + SiLU & $512$ \\
GRU and Output hidden state ($h_{t}$) & $512$ \\
\bottomrule
\end{tabular}
\label{table:sequential}
\end{table*}

\begin{table*}[!ht]
\caption{The Representation Network process encoded features $x_{t}$ and hidden state $h_{t}$ to generate the model stochastic state $z_{t}$ which is sampled from a one hot categorical distribution. We use batch normalization (BN)~\cite{ioffe2015batch} inside the representation network to ensure stable training.}
\scriptsize
\centering
\hfill \break
\begin{tabular}{cc}
\toprule
Module & Output shape \\
\midrule
Input features ($x_{t}$), Input hidden state ($h_{t})$ & $4096$, $512$ \\
Concatenate & $4608$ \\
Linear + BN + SiLU & $512$ \\
Linear and Reshape & $32 \times 32$ \\
Sample one hot and Output stochastic state ($z_{t}$) & $32 \times 32$ \\
\bottomrule
\end{tabular}
\label{table:representation}
\end{table*}

\begin{table*}[!ht]
\caption{The Dynamics Predictor learns to predict current stochastic state $z_{t}$ without being conditioned on encoded features $x_{t}$. The sequential network and dynamics predictor are used during the behavior learning phase to generate imaginary trajectories and train the actor-critic networks.}
\scriptsize
\centering
\hfill \break
\begin{tabular}{cc}
\toprule
Module & Output shape \\
\midrule
Input hidden state ($h_{t})$ & $512$ \\
Linear + LN + SiLU & $512$ \\
Linear and Reshape & $32 \times 32$ \\
Sample one hot and Output stochastic state ($\hat{z}_{t}$) & $32 \times 32$ \\
\bottomrule
\end{tabular}
\label{table:dynamics}
\end{table*}

\begin{table*}[!ht]
\caption{Networks with Multi Layer Perceptron (MLP) structure. Inputs are first flattened and concatenated along the feature dimension. Each MLP layer is followed by a layer normalization and SiLU activation except for the last layer which outputs distribution logits.}
\scriptsize
\setlength{\tabcolsep}{5pt}
\centering
\hfill \break
\begin{tabular}{cccccc}
\toprule
Network & MLP layers & Inputs & Hidden dimension & Output dimension & Output Distribution \\
\midrule
Reward predictor & 3 & $s_{t}$ & 512 & 255 & Symlog discrete \\
Continue predictor & 3 & $s_{t}$ & 512 & 1 & Bernoulli \\
Value predictor & 3 & $s_{t}$ & 512 & 255 & Symlog discrete \\
Action predictor & 3 & $x_{t}$, $s_{t-1}$ & 512 & A & Normal (continuous) / One hot (discrete)\\
Critic network & 3 & $s_{t}$ & 512 & 255 & Symlog discrete \\
Action network & 3 & $s_{t}$ & 512 & A & Normal (continuous) / One hot (discrete)\\
\bottomrule
\end{tabular}
\label{table:mlps}
\end{table*}

\newpage

\subsection{World Model Predictions}

\begin{figure*}[!ht]
        \centering
        \includegraphics[width=1.0\linewidth]{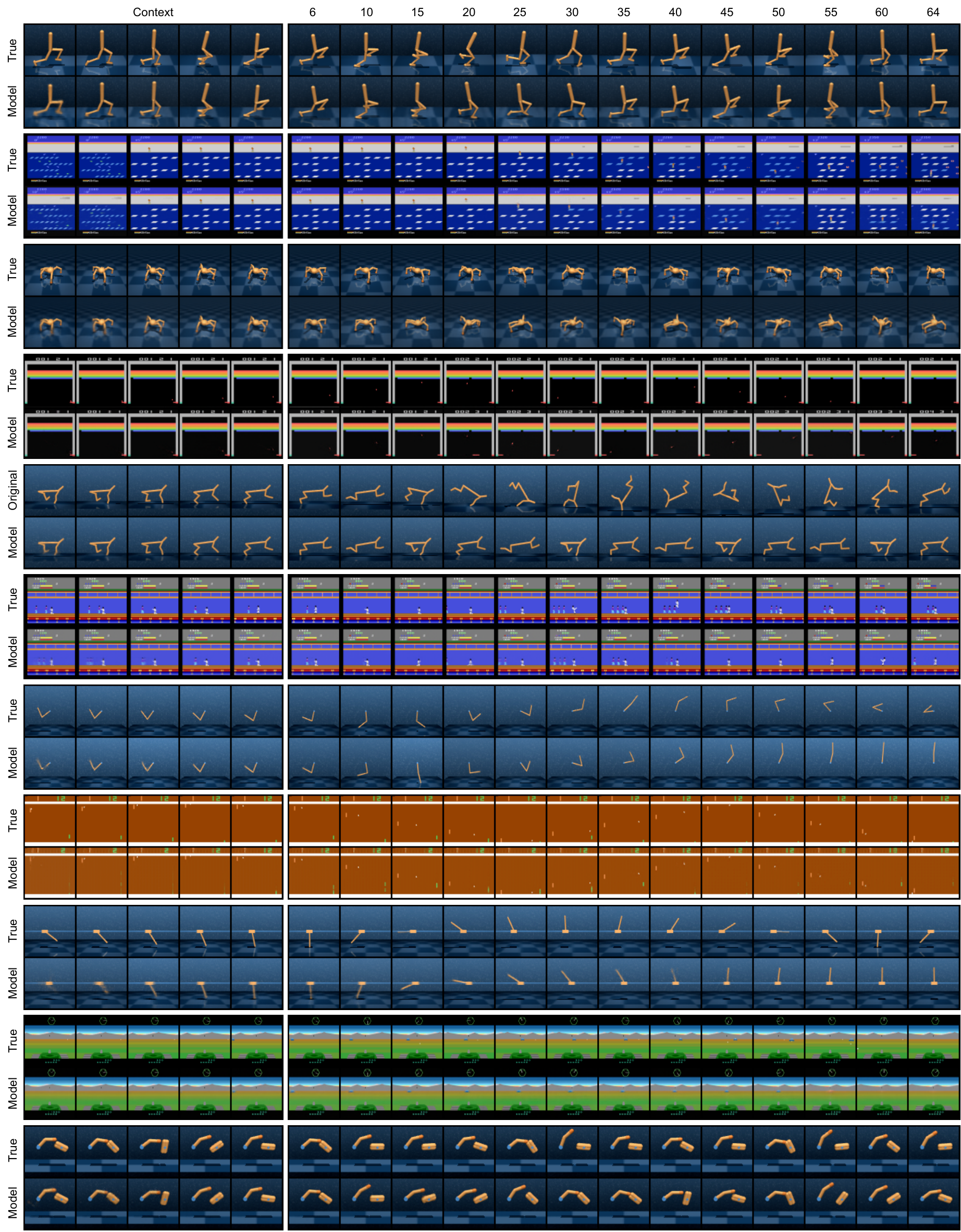}
        \caption{Trajectories imagined by the world model over 64 time steps using 5 context frames. MuDreamer generates accurate long-term predictions for various tasks without requiring reconstruction loss gradients during training to compress the observation information into the model hidden state. Although the reconstruction gradients are not propagated to the whole network, the decoder successfully reconstruct the input image, meaning that the model hidden state contains all necessary information about the environment.}
        \label{figure:sequence_nobg}
\end{figure*}

\newpage

\begin{figure*}[!ht]
        \centering
        \includegraphics[width=1.0\linewidth]{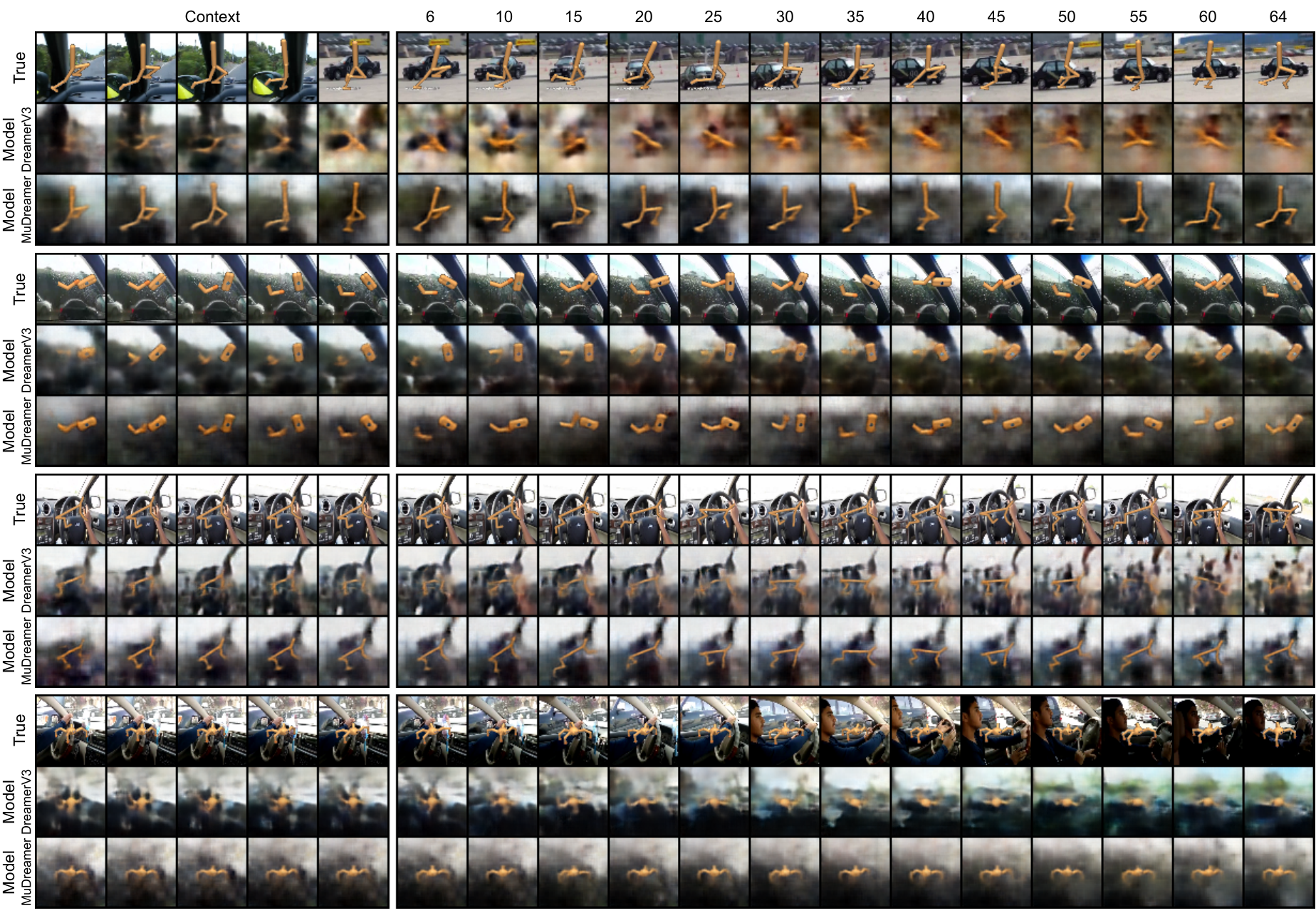}
        \caption{Trajectories imagined by DreamerV3 and MuDreamer world models over 64 time steps using 5 context frames under the natural background setting. MuDreamer models the environment dynamics in latent space, filtering the unnecessary information while DreamerV3 focuses on modeling the environment background.}
\end{figure*}

\subsection{Training Speed Comparison}

We compare MuDreamer training speed and performance with Dreamer, TPC and DreamerPro. While MuDreamer achieves similar speed at test time, the training time and memory requirement are lowered due to the non-necessity of reconstructing input observations to learn hidden representations. Table~\ref{table:training_speed} compares the training speed on the \textit{Walker-run} visual control task using a RTX 3090 GPU. We show the number of FLOPs required for a world model training forward using a sequence of 64 time frames and same architecture hyper-parameters as MuDreamer (Table~\ref{table:hyperparams}). We also experiment with larger image sizes, adding one strided convolution layer to the encoder and decoder networks when the image size is doubled. DreamerPro achieves competitive performance compared to MuDreamer but relies on the encoding of two augmented views of the same image to compute the SwAV loss. This results in a significant increase of the number of FLOPs and GPU memory, leading to slower training and memory overflow for larger image sizes. TPC benefits from good training speed since it does not requires a decoder or action-value predictors but it also has a noticeable negative impact on performance.

\begin{table*}[ht]
    \caption{Training speed comparison on \textit{Walker Run} task.} 
    \setlength{\tabcolsep}{4pt}
    \scriptsize
    \centering
    \begin{tabular}{lccccccccc}
        \toprule
          \multirow{2.5}{*}{Method} & \multicolumn{1}{c}{\multirow{2.5}{*}{\begin{tabular}[c]{@{}c@{}}Reconstruction \\ Free Model\end{tabular}}} & \multicolumn{1}{c}{\multirow{2.5}{*}{\begin{tabular}[c]{@{}c@{}}Single View \\ Training\end{tabular}}} & 
          \multicolumn{3}{c}{\#FLOPs (Billion)} & \multicolumn{3}{c}{Training step per sec} & \multicolumn{1}{c}{\multirow{2.5}{*}{\begin{tabular}[c]{@{}c@{}}Background Setting \\ mean score\end{tabular}}} \\ 
        \cmidrule(lr){4-6}\cmidrule(lr){7-9} & & & 64x64 & 128x128 & 256x256 & 64x64 & 128x128 & 256x256\\
        \midrule
        Dreamer~\cite{hafner2023mastering} & \xmark & \checkmark & 4.3 & 7.0 & 18.0 & 4.6 & 2.8 & OOM & 86.7 \\
        TPC~\cite{nguyen2021temporal} & \checkmark & \checkmark & 2.2 & 3.6 & 9.1 & 5.3 & 3.8 & 1.8 & 372.8 \\
        DreamerPro~\cite{deng2022dreamerpro} & \checkmark & \xmark & 7.9 & 12.9 & 34.9 & 3.2 & 2.0 & OOM & 445.2 \\
        MuDreamer & \checkmark & \checkmark & 2.5 & 3.8 & 9.3 & 5.0 & 3.7 & 1.8 & \textbf{517.0} \\
        \bottomrule
    \end{tabular}
    \label{table:training_speed}
\end{table*}

\newpage

\subsection{Visual Control Suite Comparison}
\label{appendix:results_dmc_1M}

\begin{figure*}[!ht]
        \centering
        \includegraphics[width=0.85\linewidth]{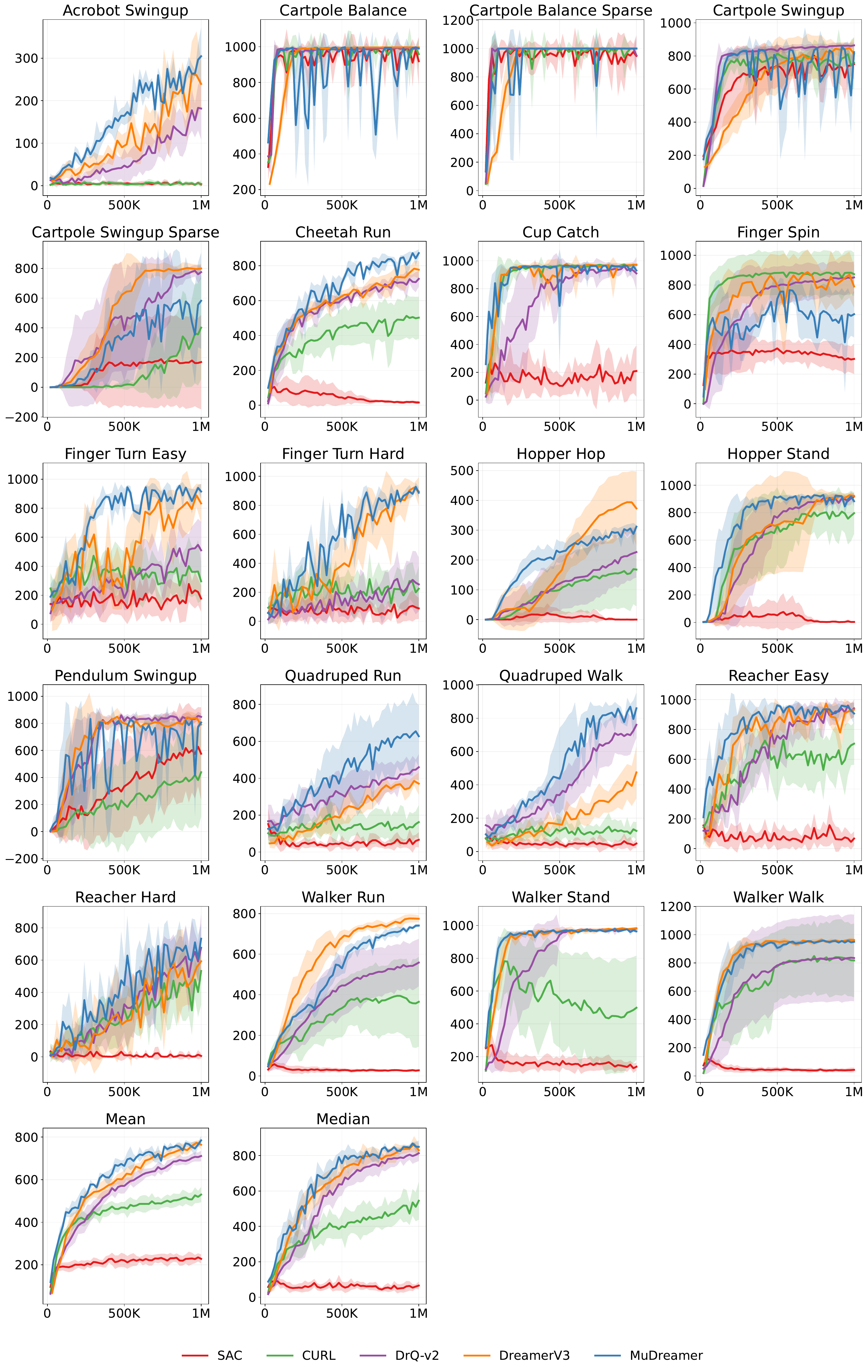}
        \caption{Comparison of MuDreamer with DreamerV3, DrQ-v2, CURL and SAC on the Deep Mind Control Suite (1M environment steps). We average the evaluation score over 10 episodes and use 3 seeds per experiment.}
        \label{figure:results_dmc_1M}
\end{figure*}

\newpage

\subsection{Natural Background Setting Comparison}

\begin{figure*}[!ht]
        \centering
        \includegraphics[width=0.85\linewidth]{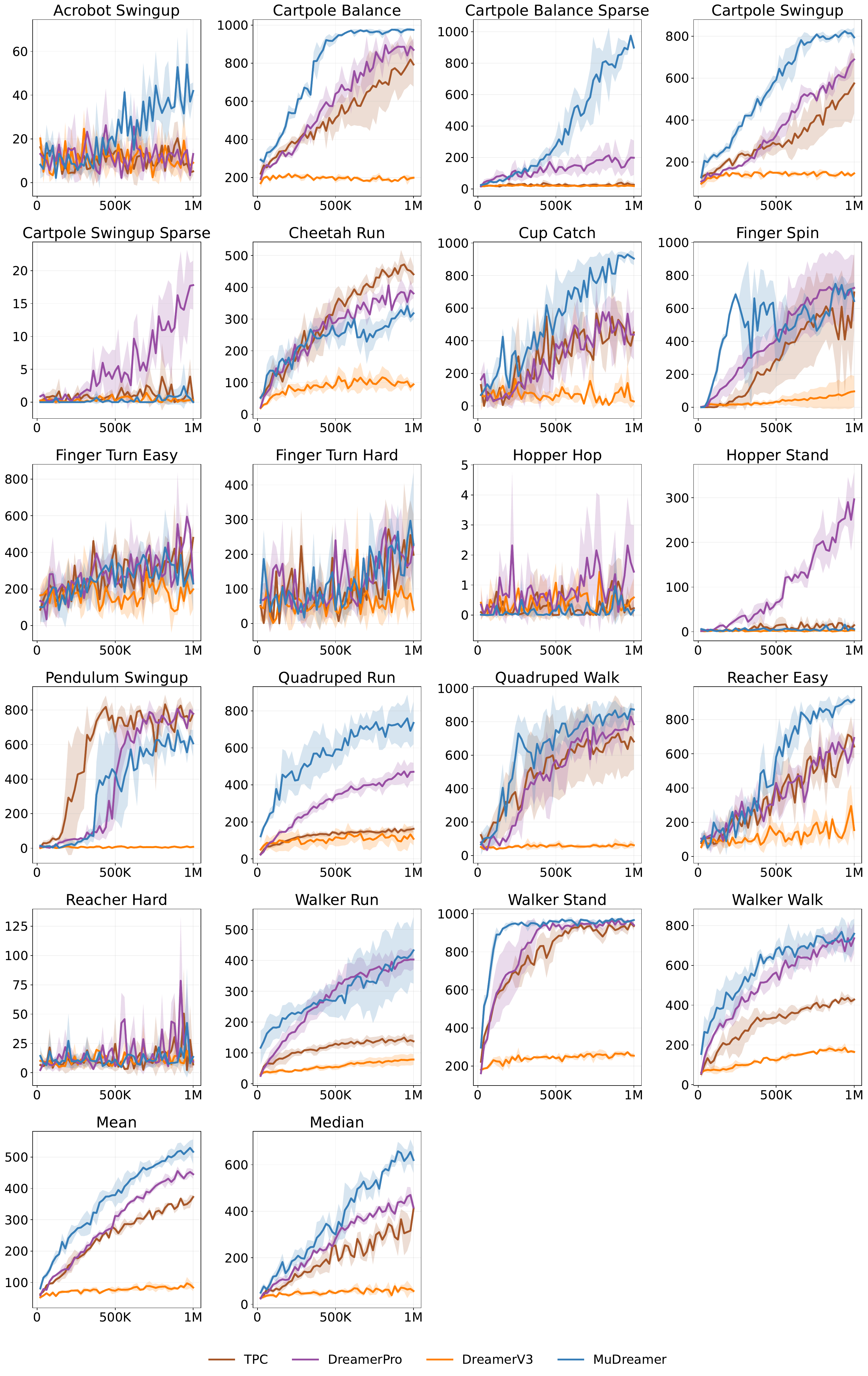}
        \caption{Comparison of MuDreamer with DreamerV3, DreamerPro and TPC under the natural background setting (1M environment steps). TPC and DreamerPro results were obtained using the official implementation of DreamerPro. We average the evaluation score over 10 episodes and use 3 seeds per experiment.}
        \label{figure:results_bg_1M}
\end{figure*}

\newpage

\subsection{Atari100k Comparison}

\begin{figure*}[!ht]
        \centering
        \includegraphics[width=0.85\linewidth]{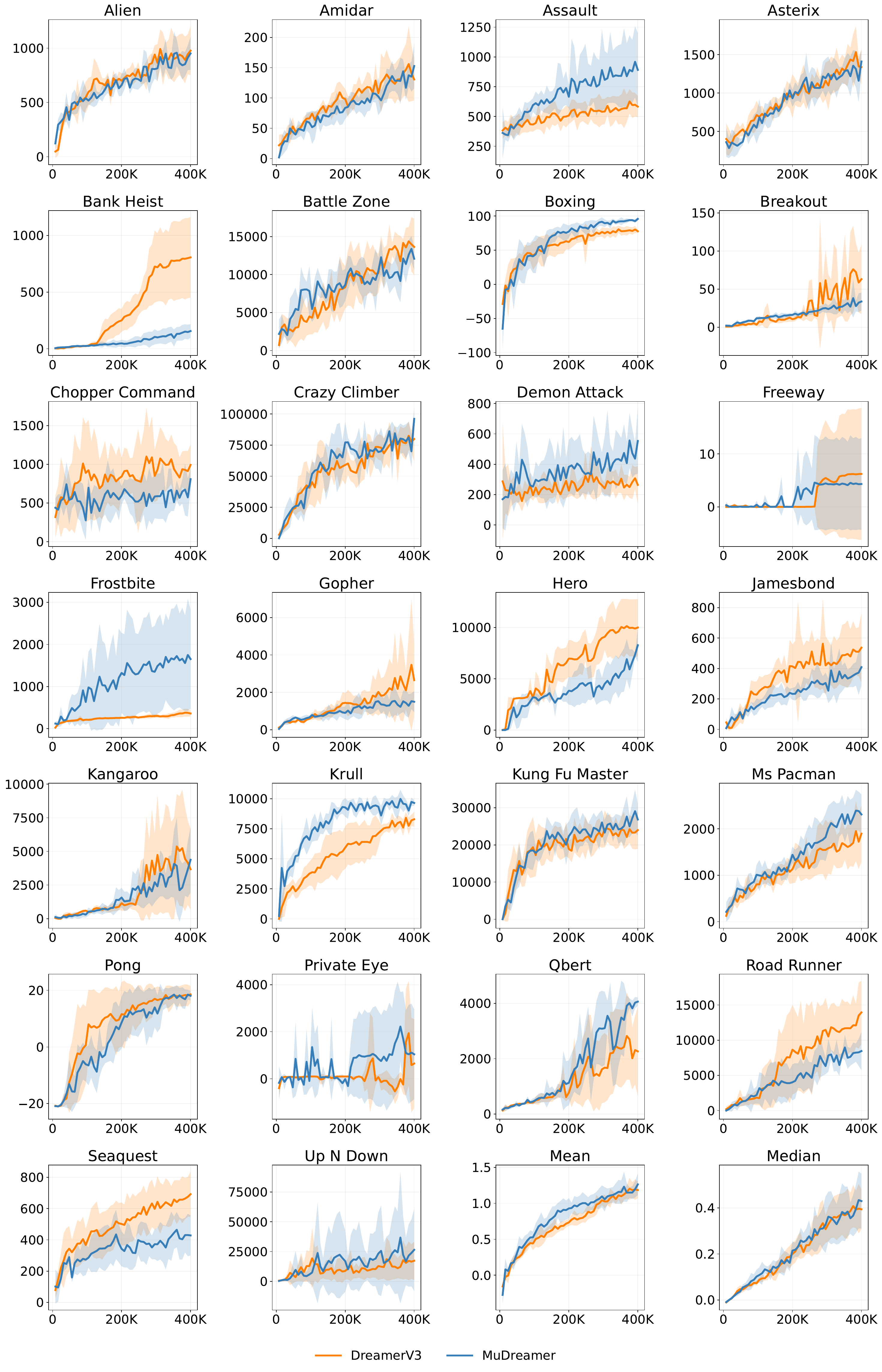}
        \caption{Comparison of MuDreamer and our tested re-implementation of the DreamerV3 algorithm on the Atari100k benchmark (400K environment steps). We average the evaluation score over 10 episodes and use 5 seeds per experiment.}
        \label{figure:results_atari100k}
\end{figure*}

\newpage

\subsection{Batch Normalization Ablation}
\label{appendix:ablation_bn}

\begin{figure*}[!ht]
        \centering
        \includegraphics[width=0.85\linewidth]{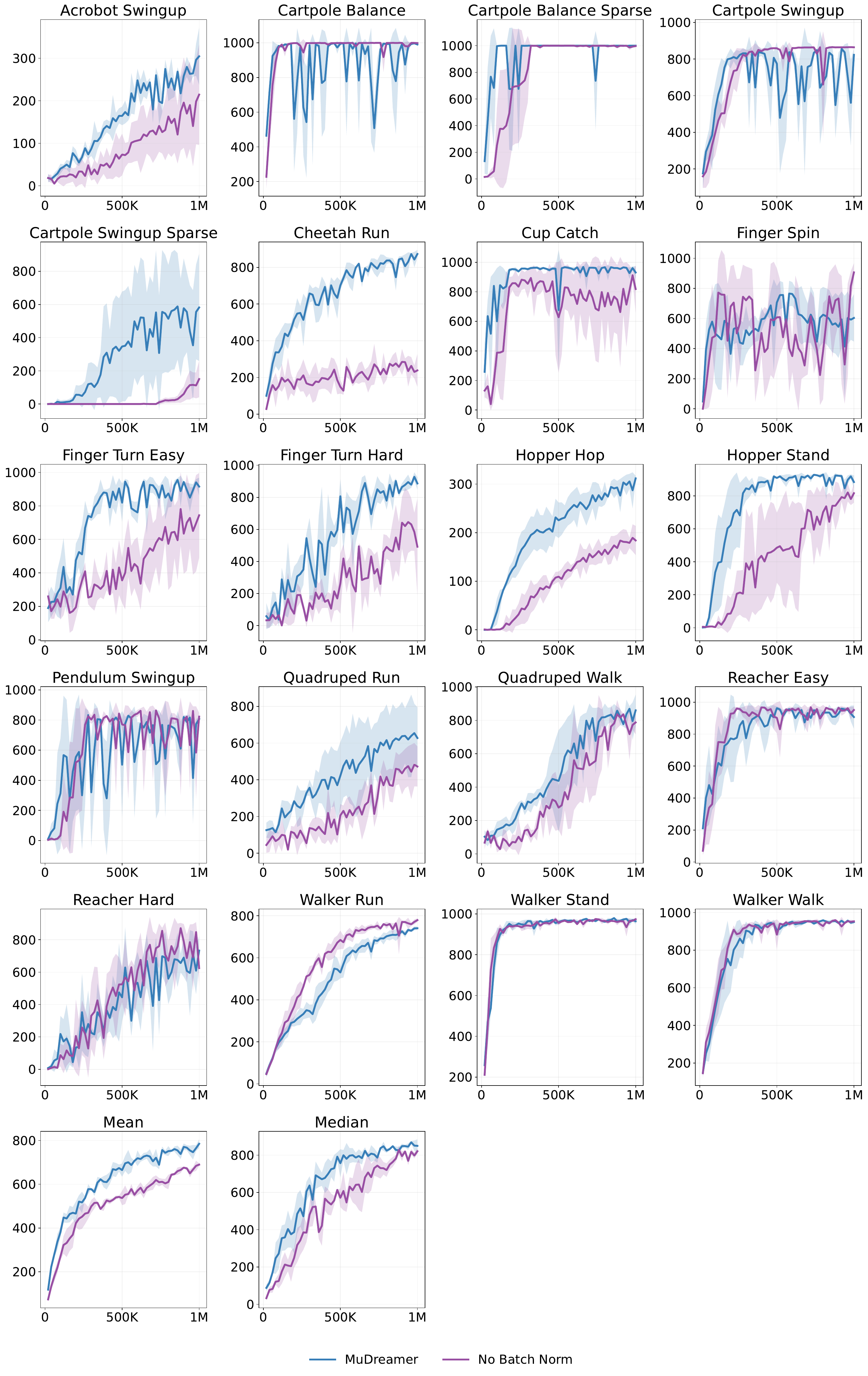}
        \caption{Comparison of MuDreamer, using layer normalization instead of batch normalization in the representation network. Removing the batch normalization layer leads to optimization difficulties for some of the tasks. The model hidden states collapse to constant or non-informative representations. This makes it impossible for the decoder to reconstruct the input observations.}
        \label{figure:ablation_bn_plots}
\end{figure*}

\newpage

\subsection{Action and Value Predictors Ablation}
\label{appendix:ablation_actor-critic}

\begin{figure*}[!ht]
        \centering
        \includegraphics[width=0.85\linewidth]{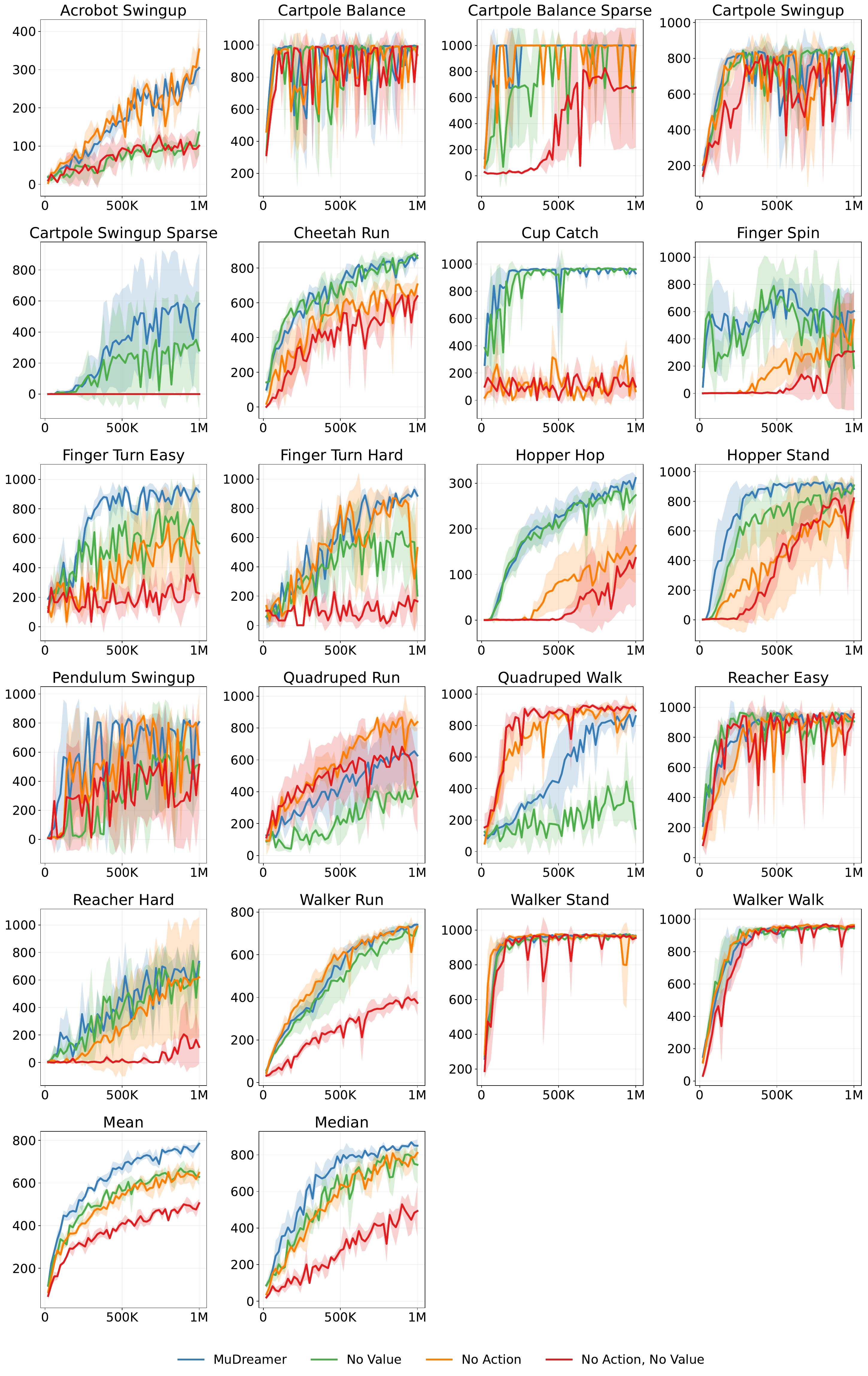}
        \caption{Comparison of MuDreamer, removing the action and/or value prediction heads. Removing the action and/or value heads during training deteriorates MuDreamer performance. We find that action prediction significantly helps tasks with sparse rewards like Hopper Hop or Cartpole Swingup Sparse. The value prediction head improves stability, leading to more stable learning. Both heads help the model to learn representations, leading to faster convergence and lower reconstruction loss.}
        \label{figure:ablation_action-value_plots}
\end{figure*}

\newpage

\subsection{KL Balancing Ablation}
\label{appendix:ablation_kl}

\begin{figure*}[!ht]
        \centering
        \includegraphics[width=0.85\linewidth]{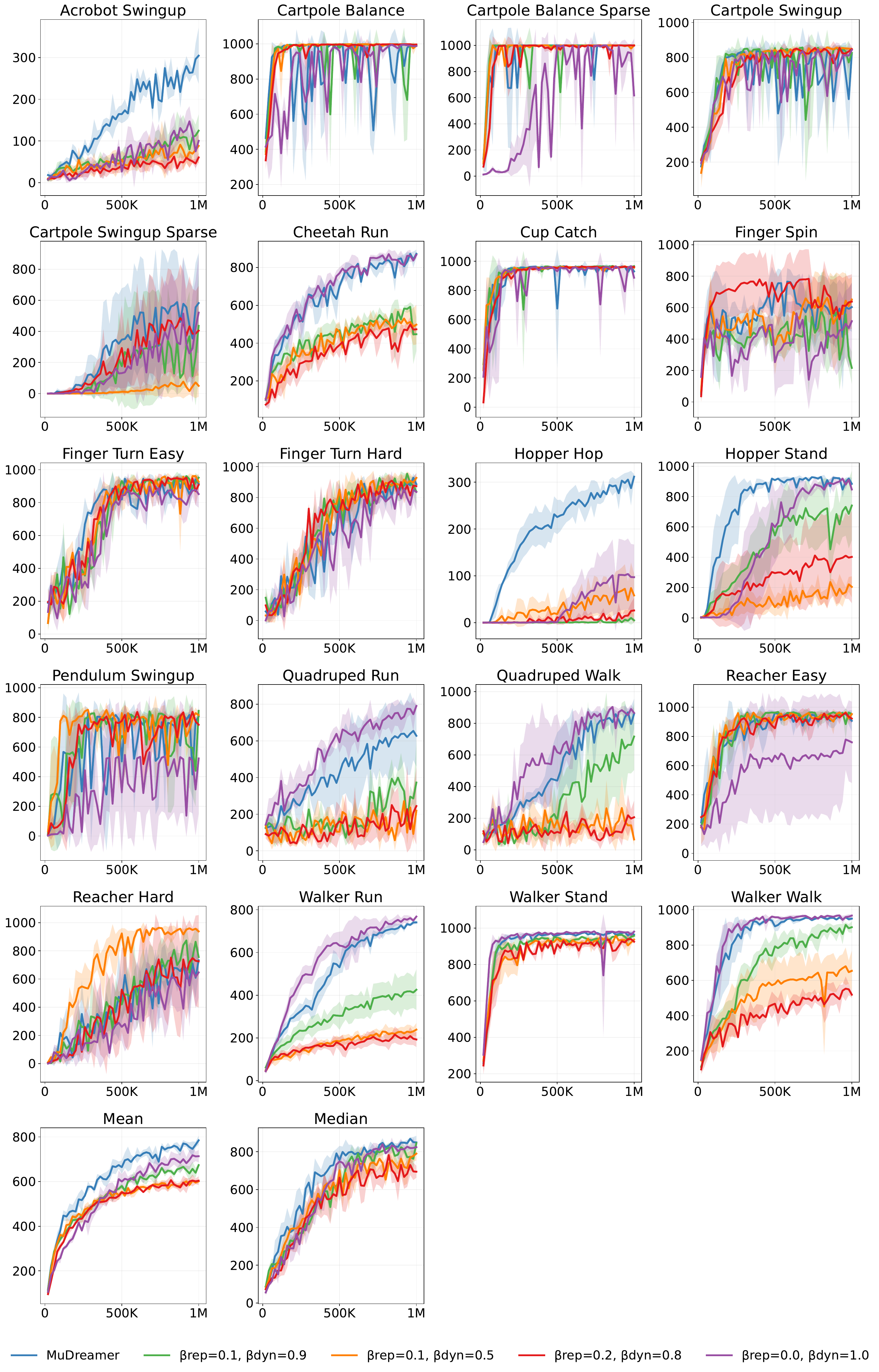}
        \caption{Comparison of MuDreamer, studying the effect of $\beta_{dyn}$ and $\beta_{rep}$. We find that applying the default KL balancing parameters of DreamerV3 slows down convergence for some of the tasks, restraining the world model from learning representations. Setting $\beta_{rep}$ to zero improves learning speed but results in instabilities and a degradation of performance after a certain amount of steps when training longer.
         Using a slight regularization of the representations toward the prior with $\beta_{rep} = 0.05$ improved convergence speed while maintaining stability.}
        \label{figure:ablation_kl_plots}
\end{figure*}

\newpage

\subsection{Additional Visualizations}

\begin{figure*}[!ht]
        \centering
        \includegraphics[width=0.9\linewidth]{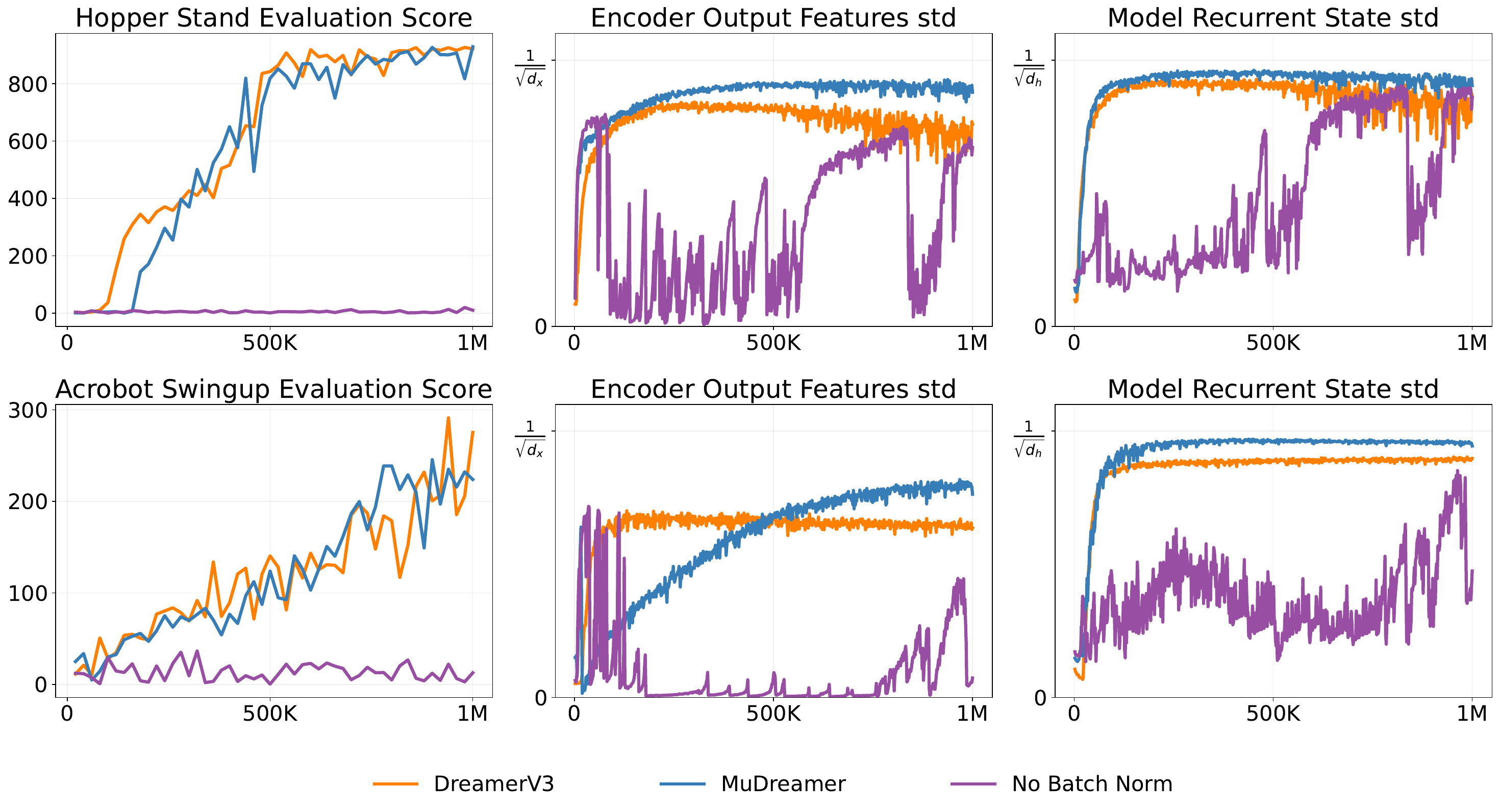}
        \caption{DreamerV3 and MuDreamer comparison with or without using batch normalization in the representation network instead of layer normalization (train ratio = 128). Left plots: agent evaluation score. Middle: the per-channel std of the l2-normalized encoder outputs $x_{t}$, plotted as the averaged std over all channels. Right plots: per-channel std of the l2-normalized model recurrent state $h_{t}$. Batch normalization improves learning stability and prevents collapses to constant feature vectors for all images.}
\end{figure*}

\begin{figure}[!ht]
        \centering
        \includegraphics[width=0.9\linewidth]{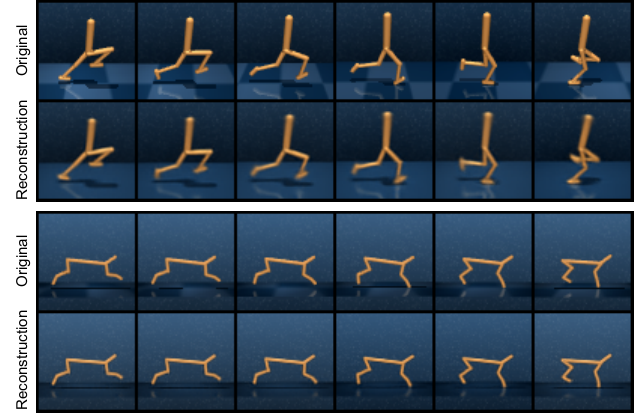}
        \caption{Example of MuDreamer reconstruction on Walker Run and Cheetah Run tasks using default KL balancing parameters of DreamerV3 ($\beta_{dyn}=0.5$, $\beta_{rep}=0.1$). The default regularization of the posterior representations toward the prior tends to limit the amount of information in the latent space in order to minimize the KL divergence between the two state distributions. In this examples, we observe that unnecessary information such as the environment floor are reconstructed as monochrome surfaces without the original details.}
\end{figure}

\newpage

\subsection{Natural Background Setting 6 tasks Comparison}

\begin{figure*}[!ht]
        \centering
        \includegraphics[width=0.9\linewidth]{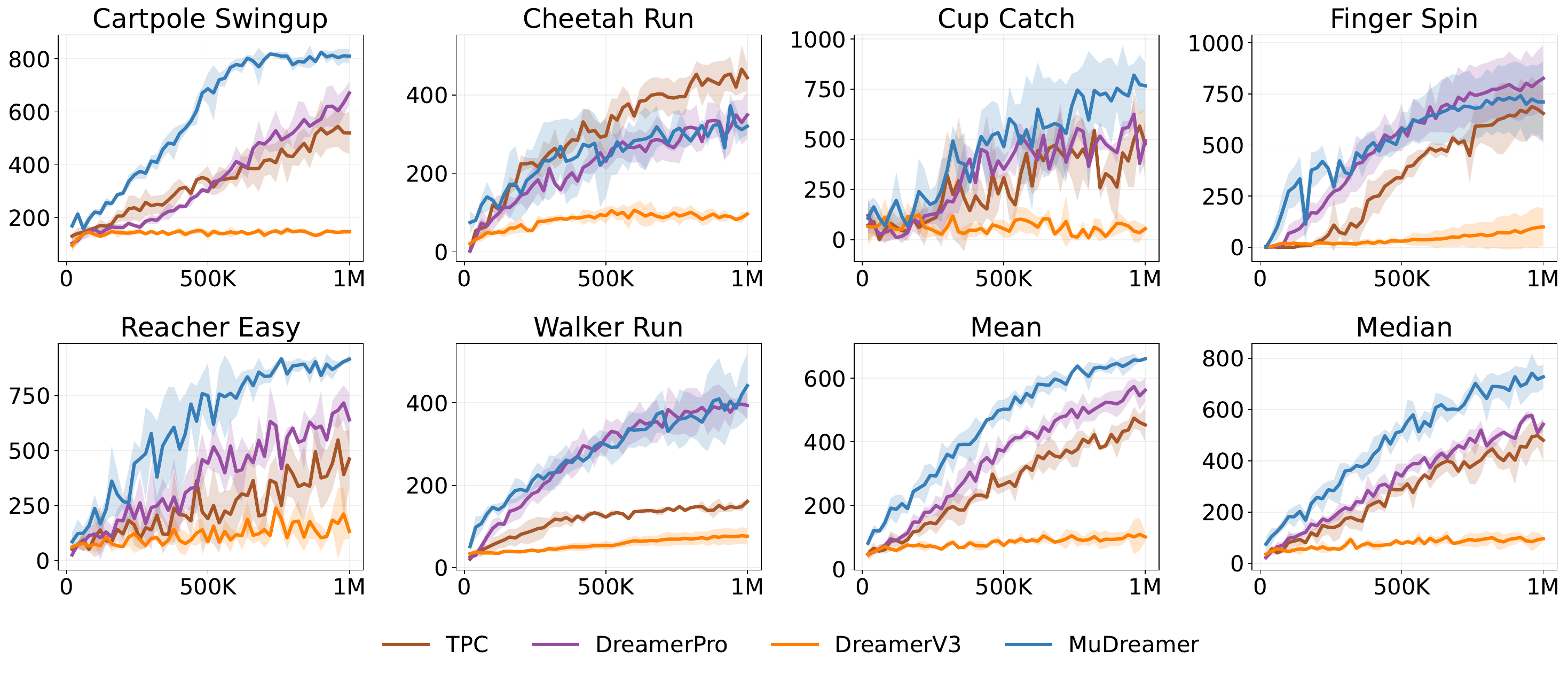}
        \caption{Comparison of MuDreamer with DreamerV3, DreamerPro and TPC on the six visual control tasks considered in~\citet{deng2022dreamerpro} under the natural background setting (1M env. steps).}
        \label{figure:results_dmc_bg_6tasks}
\end{figure*}

We compare MuDreamer with DreamerPro and TPC on the six visual control tasks considered in~\citet{deng2022dreamerpro}. Following DreamerPro, we scale the reward prediction loss by a factor of 100 to further encourage extraction of task-relevant information. Figure~\ref{figure:results_dmc_bg_6tasks} shows that MuDreamer outperforms DreamerPro and TPC over the six tasks without requiring encoding separate augmented views of input images during training. MuDreamer achieves a mean score of 661.3 over the six tasks against 562.4 and 453.3 for DreamerPro and TPC.

\begin{table}[!ht]
    \caption{Visual Control Suite scores on the six tasks considered in~\citet{deng2022dreamerpro} under the natural background setting (1M environment steps). $\dagger$ denotes our tested reimplementation of DreamerV3. $\ddagger$ results were taken from DreamerPro. We average the evaluation score over 10 episodes and use 3 seeds per experiment.} 
    \setlength{\tabcolsep}{10pt}
    \scriptsize
    \centering
    \hfill \break
    \begin{tabular}{lcccc}
    \toprule
    Task & TPC$^{\ddagger}$ & DreamerPro$^{\ddagger}$ & DreamerV3$^{\dagger}$ & MuDreamer \\ 
    \midrule
    Cartpole Swingup & 520.7 & 671.2 & 146.8 & \textbf{810.4} \\
    Cheetah Run & \textbf{444.2} & 349.1 & 96.5 & 320.1 \\
    Cup Catch & 477.2 & 493.4 & 54.6 & \textbf{767.6} \\
    Finger Spin & 654.8 & \textbf{826.4} & 99.8 & 711.5 \\
    Reacher Easy & 462.1 & 640.6 & 132.3 & \textbf{916.7} \\
    Walker Run & 160.7 & 393.8 & 76.5 & \textbf{441.6} \\
    \midrule
    Mean & 453.3 & 562.4 & 101.1 & \textbf{661.3} \\
    Median & 480.1 & 542.1 & 96.3 & \textbf{727.2} \\
    \bottomrule
    \end{tabular}
    \label{table:results_dmc_bg_1M_6tasks}
\end{table}

%%%%%%%%%%%%%%%%%%%%%%%%%%%%%%%%%%%%%%%%%%%%%%%%%%%%%%%%%%%%

\end{document}